\pdfoutput=1

\documentclass[11pt]{article}

\usepackage{EMNLP2023}

\usepackage{times}
\usepackage{latexsym}

\usepackage[T1]{fontenc}

\usepackage[utf8]{inputenc}

\usepackage{microtype}

\usepackage{inconsolata}

\usepackage{pifont}
\usepackage{makecell}
\usepackage{multirow} 
\usepackage{amsmath}
\usepackage{subcaption}
\usepackage{graphicx}
\usepackage{booktabs}
\usepackage{xspace}
\newcommand{\squeezespaces}[1]{%
  \thickmuskip=#1\thickmuskip
  \medmuskip=#1\medmuskip
  \thinmuskip=#1\thinmuskip
  \nulldelimiterspace=#1\nulldelimiterspace
  \scriptspace=#1\scriptspace
}

\usepackage{xcolor}
\usepackage{xspace}
\definecolor{purple(munsell)}{rgb}{0.62, 0.0, 0.77}

\newcolumntype{L}[1]{>{\raggedright\let\newline\\\arraybackslash\hspace{0pt}}p{#1}}
\newcolumntype{M}[1]{>{\raggedright\let\newline\\\arraybackslash\hspace{0pt}}m{#1}}
\newcolumntype{C}[1]{>{\centering\arraybackslash\hspace{0pt}}p{#1}}

\newcommand{\our}{\textsc{BTW}\xspace}

\title{\our: A Non-Parametric Variance Stabilization Framework for Multimodal Model Integration}

\author{
  Jun Hou \textsuperscript{$\spadesuit$},
  Le Wang \textsuperscript{$\diamondsuit$},
  Xuan Wang \textsuperscript{$\spadesuit$}\\
  \textsuperscript{$\spadesuit$} Department of Computer Science, Virginia Tech, Blacksburg, VA, USA\\
  \textsuperscript{$\diamondsuit$} Department of Agricultural and Applied Economics, Virginia Tech, Blacksburg, VA, USA\\
  \texttt{\{junh, lewangecon, xuanw\}@vt.edu}
}

\begin{document}
\maketitle
\begin{abstract}

Mixture-of-Experts (MoE) models have become increasingly powerful in multimodal learning by enabling modular specialization across modalities. However, their effectiveness remains unclear when additional modalities introduce more noise than complementary information. Existing approaches, like the Partial Information Decomposition, struggle to scale beyond two modalities and lack instance-level control.
We propose \textbf{B}eyond \textbf{T}wo-modality \textbf{W}eighting (\our)~\footnote{https://github.com/JuneHou/Multimodal-Infomax-moe.git}, a bi-level, non-parametric weighting framework that combines instance-level Kullback-Leibler (KL) divergence and modality-level mutual information (MI) to dynamically adjust modality importance during training. Our method requires no extra parameters and supports any number of modalities. Specifically, \our computes per‐example KL weights by measuring the divergence between each unimodal and the current multimodal prediction, and modality‐wide MI weights by estimating global alignment between unimodal and multimodal outputs. Extensive experiments on sentiment regression and clinical classification demonstrate that our method significantly improves regression performance and multiclass classification accuracy.

\end{abstract}



\section{Introduction}
\begin{figure}[ht]
    \centering
    \includegraphics[width=0.48\textwidth]{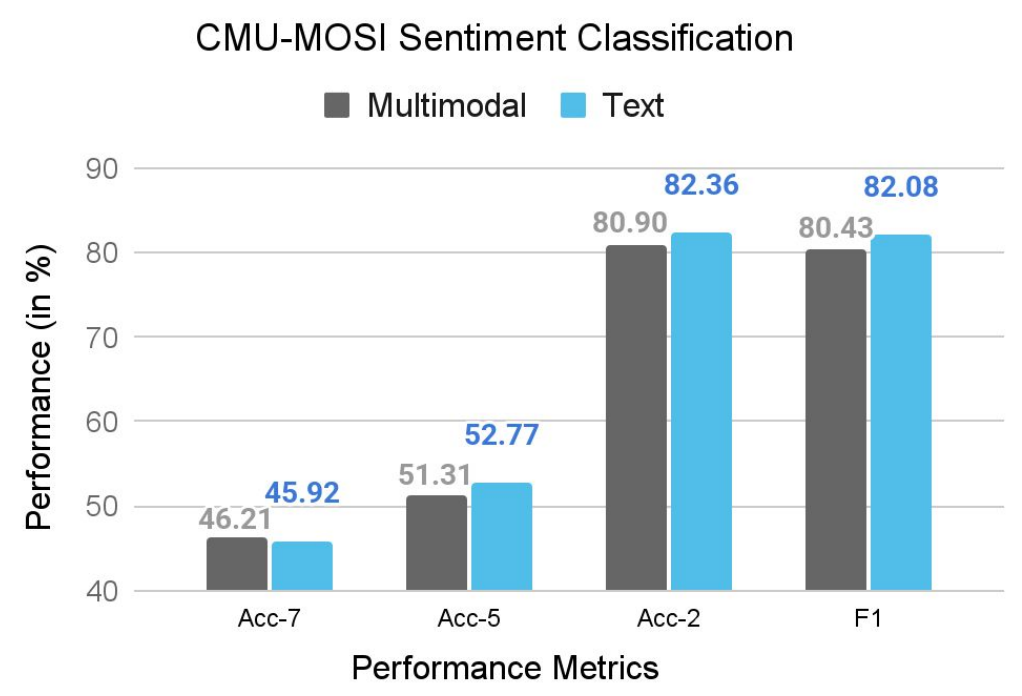}
    \caption{Illustration of our motivation. The CMU-MOSI dataset text modality stand alone could performs better than the multimodal in 5-class classification (Acc-5), binary classification (Acc-2) and Weighted-F1 score. }
    \label{fig:text_better}
\end{figure}
Multimodal learning has advanced rapidly in vision-language reasoning~\cite{lin2024moe}, emotion recognition~\cite{zadeh2018multimodal}, and clinical decision support~\cite{soenksen2022integrated}. Sparsely-gated Mixture-of-Experts (MoE) models have emerged as powerful and efficient solutions for scaling multimodal architectures through modular specialization across expert subnetworks~\cite{shazeer2017outrageously, fedus2022switch}. Designed for modality-specific and cross-modal patterns, MoE models have achieved success in multimodal fusion, including vision-language grounding, representation learning, and alignment~\cite{Feng2022ERNIEViLG2IA, mustafa2022multimodal}. Recent MoE frameworks flexibly integrate three or more modalities, achieving strong results across domains~\cite{han2024fusemoe, NEURIPS2024_b2f2af54, li2025uni}. 

Despite these advances, it is unclear if integrating multiple modalities adds more noise than useful information, or if current MoE designs capture cross-modal interactions effectively. As shown in Figure~\ref{fig:text_better}, the text-only model outperforms the full multimodal system in 5-class and binary sentiment classification (Acc-5, Acc-2), as well as Weighted-F1, on the CMU-MOSI dataset \cite{zadeh2016mosi}. This result suggests that merely aggregating modalities can hurt performance when one view is substantially more informative, as also seen in healthcare \cite{hager2024evaluation}, highlighting the need for a mechanism that both stabilizes variance and selectively amplifies complementary information to achieve a more effective multimodal integration.
\begin{table}[t]
\centering
\small
\caption{Comparison of modality interaction methods by scalability, parametric nature, task-specificity, and bi-level design. BTW uniquely supports all four, enabling robust and efficient modeling.}
\resizebox{\linewidth}{!}{%
\begin{tabular}{lcccc}
\hline
 & \makecell{\textbf{Scalable}} 
 & \makecell{\textbf{Non-} \\ \textbf{parametric}} 
 & \makecell{\textbf{Task-}\\ \textbf{agnostic}} 
 & \makecell{\textbf{Bi-level}} \\
\hline
\textbf{\our}          & \ding{52} & \ding{52} & \ding{52} & \ding{52} \\
\textbf{MI}          & \ding{52} & \ding{52} & \ding{52} & \ding{56} \\
\textbf{PID}           & \ding{56} & \ding{52} & \ding{56} & \ding{56} \\
\textbf{Game-theory}   & \ding{52} & \ding{56} & \ding{52} & \ding{56} \\
\textbf{Attention}     & \ding{52} & \ding{56} & \ding{52} & \ding{56} \\
\hline
\end{tabular}
}
\label{tab:methods_compare}
\end{table}
Although methods for modeling modality interactions as solutions exist (Table~\ref{tab:methods_compare}), current approaches remain limited. Mutual information (MI)-based methods~\cite{shannon1948mathematical, han2021improving, he2024efficient} select informative modalities, but lack fine-grained variance control. Partial Information Decomposition (PID) \cite{williams2010nonnegative, liang2023quantifying} provides a principled way to disentangle information among modalities, but is hard to scale up and to generalize to regression tasks. Game-theoretic frameworks \cite{kontras2024multimodal} and attention mechanisms \cite{zhang2023learning} learn dynamic modality weights but are computationally costly. 

To address these limitations, we propose \our, a novel bi-level weighting mechanism that stabilizes variance across modalities and manages noise from added modalities. At the instance-level, \our employs Kullback-Leibler (KL) divergence \cite{kullback1997information} to measure how much multimodal predictions capture the distributional information from each individual modality. At the modality-level, it leverages MI to quantify global reliability and contribution of each modality's prediction across the dataset. By combining KL and MI, \our dynamically balances instance-level contributions with global modality reliability, reducing variance while preserving complementary information. Based on experimental results, our framework improves model stability and performance across both continuous (regression) and categorical (classification) tasks, as demonstrated by performance gains in emotion recognition tasks (\cite{zadeh2016mosi}, \cite{bagher-zadeh-etal-2018-multimodal}) and clinical length-of-stay prediction tasks \cite{johnson2023mimic}. 

\section{Related Work}
\subsection{Multimodal Fusion with MoE} 
Recent multimodal MoE works enhanced fine-grained modality understanding and cross-modal interaction through the use of modality-aware experts~\cite{lin2024moma}, local-to-global expert hierarchies~\cite{cao2023multi}, and text-guided expert activation~\cite{zhao2024tgmoe}. Others methods are designed so that different modalities selectively guide expert usage, such as the sparsely activated architecture in SkillNet~\cite{dai2022one} and the gating-based dynamic fusion in DynMM~\cite{xue2023dynamic}. These methods aim to dynamically match modality complexity with appropriate expert pathways. To further scale MoE architectures beyond two modalities and for large-scale tasks, recent approaches integrate modality-specific encoders and expert parallelism into unified models. For instance, FuseMoE~\cite{han2024fusemoe} introduces \textit{per-modality} and \textit{disjoint} routers to handle heterogeneous modality integration. Flex-MoE~\cite{NEURIPS2024_b2f2af54} supports scalable integration of any subset of modalities through a unified, flexible expert routing mechanism. Uni-MoE~\cite{li2025uni} decouples data and model parallelism across modality-specific experts. IMP~\cite{akbari2023alternating} leverages alternating gradient descent to integrate multimodal perception efficiently.

\subsection{Modality Interaction Modeling} \label{sec: interaction}

\begin{figure*}[h]
    \centering
    \includegraphics[width=0.95\textwidth]{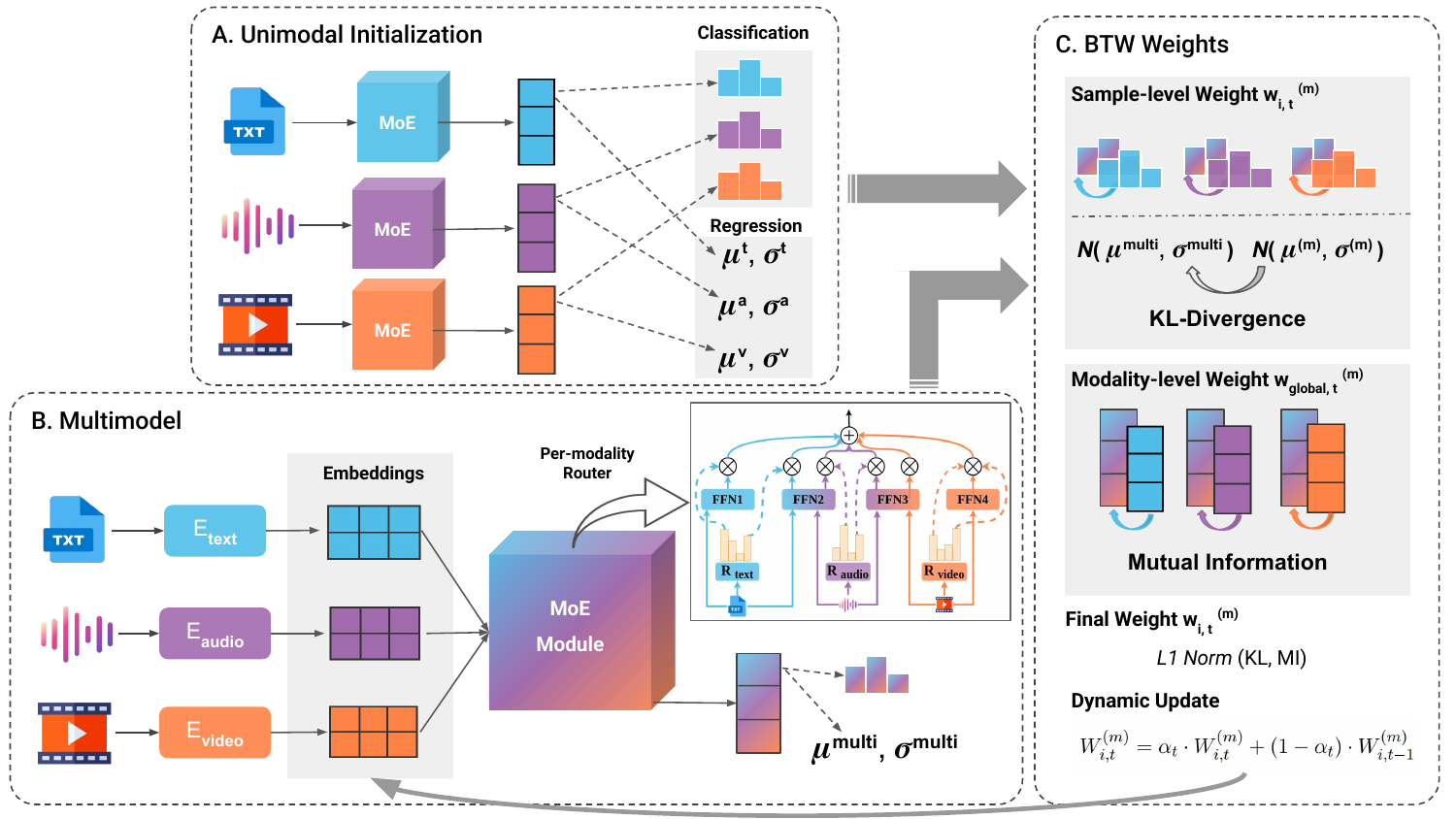}
    \caption{The overall architecture of the proposed \our weighting framework. (A) Unimodal Initialization: Each modality is processed separately through the shared MoE backbone to produce unimodal predictions $\hat{y}^{(m)}_i$, yielding a Gaussian distribution ($\mu^{(m)}_i$, $\sigma^{(m)}_i$) for regression and categorical probabilities for classification. (B) Multimodal: All embeddings of each modality are fused by a MoE module with the per-modality routers, producing a prediction ($\mu^{multi}_i$, $\sigma^{multi}_i$) for regression or categorical probabilities for classification. (C) \our Weights: The instance-level and modality-level weights are computed based on the predictions from (A) and (B). The final L1-normalized weights $W^{(m)}_i$ are dynamically smoothed across epochs to rescale modality embeddings during training.}
    \label{fig:main}
\end{figure*}

Before modality interactions can be effectively modeled, the missingness handling \cite{lin2023missmodal}, contrastive learning \cite{poklukar2022geometric} and data augmentation \cite{lin2024adapt} have improved robustness in multimodal representations. Non-statistical approaches interpret or balance modality contributions using attention-based fusion~\cite{tsai-etal-2019-multimodal, zhang2023learning} and gradient-based visualization~\cite{chen2023multiviz}. Statistical interaction modeling is favored for its model-agnostic, non-parametric nature and is applied at either global or instance levels.

At the global level, MI has been used to model modality interactions~\cite{han2021improving, he2024efficient}, and the information bottleneck has been used to reduce noise~\cite{wu2023denoising}. But these lack instance-level or directional specificity, limiting use for heterogeneous data. PID \cite{williams2010nonnegative, liang2023quantifying} decomposes information into unique and shared contributions from each modality but is typically limited to two modalities by computational cost. Game-theoretic frameworks~\cite{kontras2024multimodal} further leverage mutual information decomposition to balance modality influence across the dataset. Although these frameworks are generalizable to high-dimensional settings, the assumption of modality competition tends to overlook the need to resolve conflicts between modalities. At the instance level, DIME~\cite{lyu2022dime} attributes model predictions to each modality for individual samples, though its computational complexity can be a limitation. Recent work uses information bottleneck for instance-level modality contribution~\cite{fang2024dynamic}, filtering noise rather than modeling interactions. 


\section{Method} \label{sec:method}
\vspace{-1mm}
Our proposed \our framework, as shown in Figure~\ref{fig:main}, specifically addresses the need for a generalizable solution capable of scaling to arbitrary modalities by systematically analyzing modality interactions while simultaneously stabilizing variance. The framework is built on top of existing MoE models \cite{han2024fusemoe} and involves three steps: (1) obtaining unimodal predictions, (2) computing bi-level weights, and (3) dynamically applying these weights to modality embeddings.

\subsection{Unimodal Predictions Initialization}

To begin, we aim to extract the maximum amount of information from the input embeddings of each individual modality \( X_i^{(m)} \) in the complete data \( X_i \), where \( m \in \{1, \ldots, M\} \) and \( M \) is the total number of modalities. Consider a multimodal classifier \( f \) capable of handling an arbitrary number of modalities. To establish a baseline measure of information provided by each individual modality, we first generate unimodal predictions $\hat{y}_i^{(m)}$ by training \( f \) using only one modality at a time:
\begin{align}
    \hat{y}_i^{(0)} &= f^{(0)}(X_i^{(0)}) \\
    \hat{y}_i^{(1)} &= f^{(1)}(X_i^{(1)}) \\
    &\vdots \notag \\
    \hat{y}_i^{(M)} &= f^{(M)}(X_i^{(M)}).
\end{align}

Next, we train the same model using all available modalities jointly with existing MoE models, without applying any modality-specific weighting:
\begin{align}
    \hat{y}_i^{\text{multi}} &= f^{\text{multi}}(X_i).
\end{align}

After training the unimodal and multimodal, we collect the model prediction \(\hat{y}^{(m)}\) and \(\hat{y}^\text{multi}\). These unimodal and multimodal predictions serve as essential reference points for quantifying the individual contributions and variability of each modality in subsequent training phases.


\subsection{\our: Instance-Level Weights}

Instance-level weights are computed based on KL divergence between the probability distributions of unimodal and multimodal predictions.

\paragraph{Classification} tasks can directly output the probability, which for each modality and instance. For mordality $m$, the instance-level weight is calculated as the KL-divergence  between the unimodal and multimodal prediction distributions: 
\begin{equation}
w^{(m)}_{i} = D_{KL} (P(\hat{y}_i^{(m)}|X_i^{(m)})\;\|\; P(\hat{y}_i^\text{multi}|X_i^\text{multi})).
\label{eq:class-kl}
\end{equation}


\paragraph{Regression} tasks have continuous output from each modalities prediction. First, unimodal predictions, we model each modality’s prediction as a Gaussian distribution with mean $\mu^{(m)}_i$ and variance $\sigma^{(m)}_i$. The network output is interpreted as the conditional mean, $\mu^{(m)}_i = \mathrm{E}[\hat{y}_i^{(m)}\!\mid\!X^{(m)}_i]$. In addition, we use the squared error $\sigma_i^{(m)} = \bigl(Y_i-\mu^{(m)}_i\bigr)^2$ as an estimator of the conditional variance, where \(Y_i\) is the ground truth. Second, for multimodal predictions, \(\mu_i^{\mathrm{multi}}\) and \(\sigma_i^{\mathrm{multi}}\) are estimated using the same procedure. The unimodal estimations and the multimodal estimations are used in the closed-form KL divergence between the unimodal Gaussian $\mathcal{N}(\mu^{(m)}_i,\,\sigma_i^{(m)})$ and multimodal Gaussian $\mathcal{N}(\mu_i^{\mathrm{multi}},\,\sigma_i^{\mathrm{multi}})$, respectively (see Appendix~\ref{app:proof} for the full derivation). The instance-level weight is calculated as the KL-divergence between the above two Gaussian distributions:
{\squeezespaces{0.65}
\begin{equation}
  w_i^{(m)} = D_\text{KL}\bigl(
    \mathcal{N}(\mu^{(m)}_i,\,\sigma_i^{(m)})\;\|\;
    \mathcal{N}(\mu_i^{\mathrm{multi}},\,\sigma_i^{\mathrm{multi}})
  \bigr).
  \label{eq:instance-kl}
\end{equation}
}









These instance-level KL divergences are normalized across modalities for each instance to ensure comparability. A larger KL divergence indicates stronger disagreements between the unimodal and multimodal predictions, meaning more unique information can be learned from this modality, and thus a higher weight in the final integration,

\subsection{\our: Modality-Level Weights}

Modality-level weights are designed to quantify global modality reliability and informativeness. We calculate MI between unimodal and multimodal predictions across the entire dataset. Formally, let $\hat{y}^{(m)} = \{ \hat{y}^{(m)}_i \}_{i=1}^N$ and $\hat{y}^{\text{multi}} = \{ \hat{y}^{\text{multi}}_i \}_{i=1}^N$ denote the predicted outputs from the unimodal model for modality $m$ and the multimodal model, respectively. We define their mutual information as follows (see Appendix~\ref{app: MIproof} for details): 
\begin{multline}
\text{MI}(\hat{y}^{(m)}, \hat{y}^\text{multi}) = \\
\sum_{\hat{y}^{(m)}, \hat{y}^\text{multi} \in \mathcal{Y}} P(\hat{y}^{(m)}, \hat{y}^\text{multi}) \log \tfrac{P(\hat{y}^{(m)}, \hat{y}^\text{multi})}{P(\hat{y}^{(m)})P(\hat{y}^\text{multi})}.
\label{eq:mi-def}
\end{multline}
In classification tasks, both $\hat{y}^{(m)}$ and $\hat{y}^{\text{multi}}$ are discrete class predictions, each taking values in the set of all possible classes $\mathcal{Y}$, and mutual information is computed between each modality's predicted values and the multimodal prediction across the dataset\footnote{\url{https://scikit-learn.org/stable/modules/generated/sklearn.metrics.mutual_info_score.html}}. For regression tasks, where the predictions are continuous-valued scores, the summation in Eq.~\ref{eq:mi-def} is replaced with double integration over $\hat{y}^{(m)}$ and $\hat{y}^{\text{multi}}$. In the implementation mutual information is estimated using non-parametric entropy estimators based on $k$-nearest neighbor statistics\footnote{\url{https://scikit-learn.org/stable/modules/generated/sklearn.feature_selection.mutual_info_regression.html}}. In both cases, the resulting MI score captures how much predictive information the unimodal modality shares with the multimodal model.






\subsection{Dynamically Adapted Bi-level Weights}
The modality-level MI weights are used to rescale the instance-level KL divergence weights, amplifying contributions from globally informative modalities while attenuating the influence of less reliable ones. We experiment two versions of our \our framework, \textbf{BTW-local (KL)} defined as using instance-level weight only, and \textbf{BTW} defined as using bi-level weights. At the training epoch $t$, two versions of final weights are computes as:
\vspace{-1mm}
\begin{equation}
\text{BTW-local (KL):}\quad 
W^{(m)}_{i, t} = \frac{w^{(m)}_{i, t}}{\sum_{j=1}^{M} w^{(j)}_{i, t}}.
\label{eq:kl-only}\\
\end{equation}
\begin{equation}
\text{BTW:}\quad
W^{(m)}_{i,t} = \frac{w^{(m)}_{i,t} \cdot \text{MI}(\hat{y}^{(m)}, \hat{y}^\text{multi})}{\sum_{j=1}^{M}w^{(j)}_{i,t} \cdot\text{MI}(\hat{y}^{(j)}, \hat{y}^\text{multi})}.
\label{eq: bi-level}
\end{equation}
We dynamically update the computed bi-level weights throughout training epochs based on model performance improvements, measured by the F1 score for classification tasks or mean absolute error (MAE) for regression tasks. A smoothing factor ($\alpha_{t}$) with time step $t$ is adaptively adjusted, incremented if improves and otherwise decremented, avoiding rapid fluctuations. Specifically, the updated weights $W^{(m)}_{i,t}$ for modality at instance for the epoch are computed as:
\begin{equation}
W^{(m)}_{i,t} = \alpha_{t} \cdot W^{(m)}_{i,t} + (1 - \alpha_{t}) \cdot W^{(m)}_{i,t-1}.
\end{equation}

These stabilized weights are then multiplicatively applied to modality embeddings prior to each subsequent training epoch, emphasizing reliable and informative modalities while reducing variance across individual instances and modalities, thus enhancing robustness and generalization of the multimodal integration.


\section{Experiment}
\vspace{-1mm}
We conduct experiments to validate \our in stabilizing variance and evaluating robustness across modalities, tasks, and diverse domains. Details on computation, time cost, and hyper-parameters are in Appendix~\ref{app: hyper}.

\subsection{Datasets}
We conduct experiments on three publicly available benchmark datasets with more than two modalities:
\paragraph{CMU-MOSI} \cite{zadeh2016mosi} contains 2,199 clips collected from YouTube opinion videos labeled with sentiment in the range of -3 (negative) and 3 (positive).
\paragraph{CMU-MOSEI} \cite{zadeh2018multimodal}: 23,500 clips labeled from sentiment intensity. Both datasets are benchmarks for multimodal sentiment analysis with textual, acoustic and visual modalities.
\paragraph{MIMIC-IV} \cite{johnson2023mimic} is a large-scale clinical dataset containing rich multimodal patient data, including irregularly sampled time series (e.g., vital signs, lab tests), clinical notes, chest X-ray (CXR), and ECG signals. We adopt the dataset curation pipeline from FuseMoE \cite{han2024fusemoe}, in which only 25\% of samples include CXR modality and 52\% include ECG modality, while all samples contain time-series modality. For our experiments, we focus on predicting patient length-of-stay (LOS), re-framed as a 4-class classification task based on clinical grouping criteria proposed by CORe \cite{vanAken2021}. To evaluate the impact of missing modalities on our weighting framework, we construct a subset containing only instances with complete modality availability (statistics summarized in Table~\ref{tab:mimic-data}).
    

\begin{table}[t!]
\centering
\small
\begin{tabular}{lccccc}
\toprule
\textbf{LOS in days} & $\leq$3 & 3–7 & 7–14 & $>$14 & Total \\
\midrule
 Count &1,465 & 2,342 & 923 & 448 & 5,178 \\
\bottomrule
\end{tabular}
\caption{
Length-of-stay (LOS) class distribution for the MIMIC-IV dataset used in our experiments. 
Only the no-missing-modality subset is used for training and evaluation, containing 5,178 complete patient stays.
}
\label{tab:mimic-data}
\end{table}

\subsection{Experimental Setup}
\begin{table*}[ht]
\centering
\small
\setlength{\tabcolsep}{4pt}  
\caption{
Main results on CMU-MOSI and CMU-MOSEI for multimodal sentiment regression. 
BTW-global (KL), BTW-global (MI), BTW-local (KL) and BTW denote variants of our proposed bi-level weighting framework. Bold numbers indicate the best performance and underlined numbers indicate the second-best. All results are averaged over three runs with different random seeds. MMIM~\cite{han2021improving} and MulT~\cite{tsai-etal-2019-multimodal} results are reproduced from open-source code with the hyperparameters specified.
}
\label{tab:cmu-results}
\begin{tabular}{@{}lccccccc@{}}
\toprule
\textbf{} & \textbf{Method} & \textbf{MAE$\downarrow$} & \textbf{Corr} & \textbf{Acc-7} & \textbf{Acc-5} & \textbf{Acc-2} & \textbf{Weighted-F1} \\
\midrule
\multirow{5}{*}{\rotatebox[origin=c]{90}{\textbf{MOSI}}}
& MulT      & 0.989\scriptsize{$\pm$0.04} & 0.646\scriptsize{$\pm$0.03} & 33.33\scriptsize{$\pm$1.24} & 36.60\scriptsize{$\pm$1.77} & 77.05{\scriptsize{$\pm$1.21}} / 78.46{\scriptsize{$\pm$1.42}} & 77.01{\scriptsize{$\pm$1.12}} / 78.53{\scriptsize{$\pm$1.36}} \\
& MMIM      & 0.738\scriptsize{$\pm$0.01} & 0.778\scriptsize{$\pm$0.01} & 44.12\scriptsize{$\pm$1.31} & 50.29\scriptsize{$\pm$0.53} & \underline{82.51}{\scriptsize{$\pm$0.38}} / 84.45{\scriptsize{$\pm$1.07}} & 82.34{\scriptsize{$\pm$0.41}} / 84.37{\scriptsize{$\pm$0.93}} \\
& MoE       & 0.735\scriptsize{$\pm$0.00} & 0.770\scriptsize{$\pm$0.01} & 45.87\scriptsize{$\pm$0.59} & 52.28\scriptsize{$\pm$0.89} & 81.78{\scriptsize{$\pm$1.05}} / 83.99{\scriptsize{$\pm$1.07}} & 81.56{\scriptsize{$\pm$1.18}} / 83.88{\scriptsize{$\pm$1.02}} \\
 \cmidrule{2-8}
& BTW-global (KL) 
& 0.746\scriptsize{$\pm$0.011} & 0.774\scriptsize{$\pm$0.001} & 44.56\scriptsize{$\pm$1.80} & 51.65\scriptsize{$\pm$1.18} & 82.17{\scriptsize{$\pm$0.37}} / 84.00{\scriptsize{$\pm$0.27}} & 82.07{\scriptsize{$\pm$0.42}} / 83.96\scriptsize{$\pm$0.30} \\
& BTW-global (MI) 
& 0.726\scriptsize{$\pm$0.012} & 0.776\scriptsize{$\pm$0.003} & 44.51\scriptsize{$\pm$0.89} & 51.80\scriptsize{$\pm$1.10} & 82.34{\scriptsize{$\pm$1.06}} / \textbf{84.56}\scriptsize{$\pm$0.31} & \underline{82.49}{\scriptsize{$\pm$0.87}} / \textbf{84.66}\scriptsize{$\pm$0.23} \\
& BTW-local (KL)   & \textbf{0.714}\scriptsize{$\pm$0.01} & \textbf{0.786}\scriptsize{$\pm$0.01} & \underline{46.40}\scriptsize{$\pm$3.23} & \underline{53.26}\scriptsize{$\pm$3.31} & 82.46{\scriptsize{$\pm$0.97}} / \underline{84.55}{\scriptsize{$\pm$0.92}} & 82.33{\scriptsize{$\pm$0.96}} / \underline{84.50}{\scriptsize{$\pm$0.88}} \\
& BTW     & \underline{0.716}\scriptsize{$\pm$0.01} & \underline{0.781}\scriptsize{$\pm$0.01} & \textbf{47.52}\scriptsize{$\pm$0.77} & \textbf{54.28}\scriptsize{$\pm$1.43} & \textbf{82.75}{\scriptsize{$\pm$1.17}} / 84.35{\scriptsize{$\pm$0.84}} & \textbf{82.68}{\scriptsize{$\pm$1.26}} / 84.34{\scriptsize{$\pm$0.91}} \\
\midrule[0.5pt]
\midrule[0.5pt]
\multirow{5}{*}{\rotatebox[origin=c]{90}{\textbf{MOSEI}}}
& MulT      & 0.613\scriptsize{$\pm$0.01} & 0.669\scriptsize{$\pm$0.02} & 49.55\scriptsize{$\pm$0.49} & 50.93\scriptsize{$\pm$0.64} & 78.22{\scriptsize{$\pm$0.37}} / 80.36{\scriptsize{$\pm$1.40}} & 78.55{\scriptsize{$\pm$0.21}} / 80.42{\scriptsize{$\pm$0.89}} \\
& MMIM      & 0.578\scriptsize{$\pm$0.01} & \underline{0.728}\scriptsize{$\pm$0.01} & 51.03\scriptsize{$\pm$0.42} & 52.39\scriptsize{$\pm$0.54} & 81.61{\scriptsize{$\pm$2.37}} / 83.31{\scriptsize{$\pm$0.39}} & 81.23{\scriptsize{$\pm$2.67}} / 82.98{\scriptsize{$\pm$0.71}} \\
& MoE       & \underline{0.570}\scriptsize{$\pm$0.01} & 0.723\scriptsize{$\pm$0.01} & 52.17\scriptsize{$\pm$0.61} & 53.73\scriptsize{$\pm$0.64} & 80.02{\scriptsize{$\pm$3.84}} / \underline{83.41}{\scriptsize{$\pm$1.12}} & 80.53{\scriptsize{$\pm$3.34}} / \textbf{83.29}{\scriptsize{$\pm$1.11}} \\
\cmidrule{2-8}
& BTW-global (KL) 
& 0.572\scriptsize{$\pm$0.011} & 0.725\scriptsize{$\pm$0.010} & 52.24\scriptsize{$\pm$0.44} & 53.55\scriptsize{$\pm$0.49} & 80.90{\scriptsize{$\pm$3.83}} / 83.32\scriptsize{$\pm$0.69} & 81.14{\scriptsize{$\pm$3.19}} / \underline{83.10}\scriptsize{$\pm$0.61} \\
& BTW-global (MI) 
& \textbf{0.566}\scriptsize{$\pm$0.006} & \textbf{0.729}\scriptsize{$\pm$0.002} & \underline{52.34}\scriptsize{$\pm$0.59} & \underline{53.95}\scriptsize{$\pm$0.63} & 76.09{\scriptsize{$\pm$3.40}} / 81.78\scriptsize{$\pm$1.88} & 75.66{\scriptsize{$\pm$4.16}} / 81.68\scriptsize{$\pm$2.08} \\
& BTW-local (KL)   & \textbf{0.566}\scriptsize{$\pm$0.01} & 0.727\scriptsize{$\pm$0.01} & 52.32\scriptsize{$\pm$0.76} & 53.85\scriptsize{$\pm$0.73} & \textbf{83.02}{\scriptsize{$\pm$1.10}} / \textbf{83.60}{\scriptsize{$\pm$1.87}} & \textbf{82.81}{\scriptsize{$\pm$0.98}} / 83.07{\scriptsize{$\pm$2.29}} \\
& BTW     & 0.573\scriptsize{$\pm$0.01} & 0.722\scriptsize{$\pm$0.00} & \textbf{52.62}\scriptsize{$\pm$0.74} & \textbf{54.15}\scriptsize{$\pm$0.87} & \underline{81.97}{\scriptsize{$\pm$2.36}} / 81.92{\scriptsize{$\pm$0.96}} & \underline{81.50}{\scriptsize{$\pm$1.98}} / 81.22{\scriptsize{$\pm$1.36}} \\
\bottomrule
\end{tabular}
\end{table*}

\begin{table}[ht]
\centering
\small
\setlength{\tabcolsep}{5pt}
\caption{Main results on the MIMIC-IV dataset for length-of-stay classification. BTW-local (KL) and BTW denote variants of our bi-level weighting framework. Bold values indicate the best results across all methods. All metrics are averaged over three random seeds.}
\label{tab:mimic-results}
\begin{tabular}{lccc}
\toprule
\textbf{Method} & \textbf{Accuracy} & \textbf{Macro-F1} & \textbf{Weighted-F1} \\
\midrule
MulT        & 43.33\scriptsize{$\pm$2.08} & 33.33\scriptsize{$\pm$3.21} & 41.33\scriptsize{$\pm$0.58} \\
HAIM        & \textbf{46.00}\scriptsize{$\pm$0.00} & 33.00\scriptsize{$\pm$0.00} & 42.00\scriptsize{$\pm$0.00} \\
FuseMoE    & 41.33\scriptsize{$\pm$1.53} & \textbf{37.67}\scriptsize{$\pm$2.89} & 40.33\scriptsize{$\pm$2.31} \\
\midrule
BTW-local (KL)     & 43.67\scriptsize{$\pm$2.31} & \underline{37.00}\scriptsize{$\pm$1.00} & \underline{43.00}\scriptsize{$\pm$1.73} \\
BTW       & \underline{45.67}\scriptsize{$\pm$0.58} & \textbf{37.67}\scriptsize{$\pm$0.58} & \textbf{45.00}\scriptsize{$\pm$0.00} \\
\bottomrule
\end{tabular}
\end{table}
\paragraph{Baseline MoE Model} For all experiments, we adopt the MoE architecture from FuseMoE~\cite{han2024fusemoe} as the backbone multimodal fusion. We select the \textit{per-modality} router, the best-performing strategy in FuseMoE. The per-modality router distributes each modality independently to a shared pool of experts, offering a principled balance between modality-specific specialization and cross-modal integration. Implementation details and hyper-parameters are provided in Appendix~\ref{app: hyper}.

\paragraph{Weight Initialization and Adaptation} For both tasks, we follow the three-step procedure described in Section~\ref{sec:method}. After each training epoch, the computed bi-level weights are dynamically updated according to model performance (F1 score for classification, MAE for regression). A smoothing factor ($\alpha_t$) stabilizes these updates.

\paragraph{Evaluation Metrics}

Regression results of the MOSI and MOSEI datasets are evaluated using MAE. Additionally, following established sentiment benchmarks~\cite{han2021improving}, we report Pearson correlation (Corr), seven-class accuracy (Acc-7), five-class accuracy (Acc-5), binary classification accuracy (Acc-2) and F1 score computed for positive/negative and non-negative/negative classification results. Model performance for LOS classification is evaluated using overall Accuracy, Macro F1-score, and Weighted F1-score.

\section{Main Results}
\paragraph{Sentiment Analysis} Table~\ref{tab:cmu-results} summarizes evaluation results for multimodal sentiment regression across two benchmarks: CMU-MOSI and CMU-MOSEI. We compare our proposed bi-level weighting framework (BTW-local (KL) and BTW) against recent state-of-the-art multimodal fusion methods, including MulT~\cite{tsai-etal-2019-multimodal}, MMIM~\cite{han2021improving}, and standard MoE~\cite{han2024fusemoe}. Specifically, on CMU-MOSI, the BTW-local (KL) approach achieves the lowest MAE of 0.714 (-2\% improvement over MMIM and MoE) and the highest Pearson correlation of 0.786, outperforming all baseline models. Incorporating global MI weighting (BTW) yields competitive performance following BTW-local (KL) in regression metrics (MAE=0.716, Corr=0.781). Notably, BTW also achieves the best 7-class accuracy (47.52\%), 5-class accuracy (54.28\%), and the highest binary accuracy when including the zero-threshold (82.75\%). Overall, the BTW-local (KL) weights consistently improve the regression metrics, while BTW presents consistent out-performance in classification tasks. 

For the larger CMU-MOSEI dataset, BTW-local (KL) consistently shows strong regression results, achieving an MAE of 0.566, best binary classification accuracy (83.02\%/83.60\%) and weighted-F1 (82.81\% with zero-threshold included). Meanwhile, BTW is consistently leading in multi-class classification, reporting the highest accuracies in 7-class (52.62\%) and 5-class (54.15\%) settings. Although, BTW demonstrates reduced binary classification accuracy and F1 scores, its outstanding performance in multi-class scenarios highlights the complementary value of incorporating modality-level mutual information.

\paragraph{Length-of-Stay Prediction} We evaluate our proposed bi-level weighting framework on the MIMIC-IV dataset for the clinically relevant task of four-class length-of-stay (LOS) classification. For these experiments, we utilize only stays with complete modality data, thereby minimizing modality incompleteness as a confounding factor and focusing exclusively on evaluating the core modality interaction capacity of our weighting methods. Missing modality and 3-modality scenarios are explored separately in Appendix~\ref{app:missing} and Appendix~\ref{app:3mod}, respectively. Table~\ref{tab:mimic-results} summarizes classification results comparing our proposed BTW-local (KL) and BTW weighting variants against three strong baselines: the tree-based machine learning model HAIM~\cite{soenksen2022integrated}, the transformer-based fusion model MulT~\cite{tsai-etal-2019-multimodal}, and the recent FuseMoE~\cite{han2024fusemoe}. Our BTW weighting method achieves the highest Macro-F1 and significantly outperforms all other methods in Weighted-F1 (+5\% over FuseMoE, +4\% over MulT, and +3\% over HAIM). Additionally, BTW yields the second-highest Accuracy (45.67\%, close to the top-performing HAIM model at 46.00\%). The instance-level BTW-local (KL) attains the second-best Macro-F1 (37\%) and Weighted-F1 (43\%), exceeding all baselines, and achieves competitive Accuracy (43.67\%). These results highlight that our bi-level weighting approach substantially enhances multimodal classification performance, effectively balancing class-specific performance and overall accuracy compared to state-of-the-art baselines.

\section{Ablation Study}

\subsection{Weighting Mechanisms}
To better understand the individual contributions of each weighting component in our \our framework, we conducted ablation experiments on the sentiment analysis task. Specifically, for BTW-global (KL), we modify Eq.~\ref{eq:kl-only} by replacing the instance-level weight for each modality with its average across the dataset, thus using a constant modality-specific weight for all samples. For BTW-global (MI), we make an analogous change to Eq.~\ref{eq: bi-level}, using only the global MI value for each modality and omitting instance-level weights. Table~\ref{tab:cmu-results} summarizes the results of these ablations.

BTW-global (MI) consistently outperforms BTW-global (KL), highlighting the effectiveness of MI in capturing global modality alignment. 
Compared with BTW-global (KL), BTW-local (KL) demonstrates a clear advantage of instance-level weighting across nearly all metrics on both datasets, underscoring the benefits of fine-grained variance stabilization at the instance-level. 
Lastly, BTW reveals its complementary strengths by consistently outperforming the BTW-global (MI) in multiclass-classification accuracy (e.g., CMU-MOSEI Acc-7: 52.62\% vs. 52.34\%), indicating the synergy between local variance management and global modality informativeness. 

In summary, KL provides effective instance-level variance control beneficial to regression, while MI contributes significantly to global alignment and multi-class accuracy. The bi-level weights yields balanced, robust multimodal fusion.

\subsection{Weight Distribution Analysis}
\begin{figure}[t]
    \centering
    \includegraphics[width=0.48\textwidth]{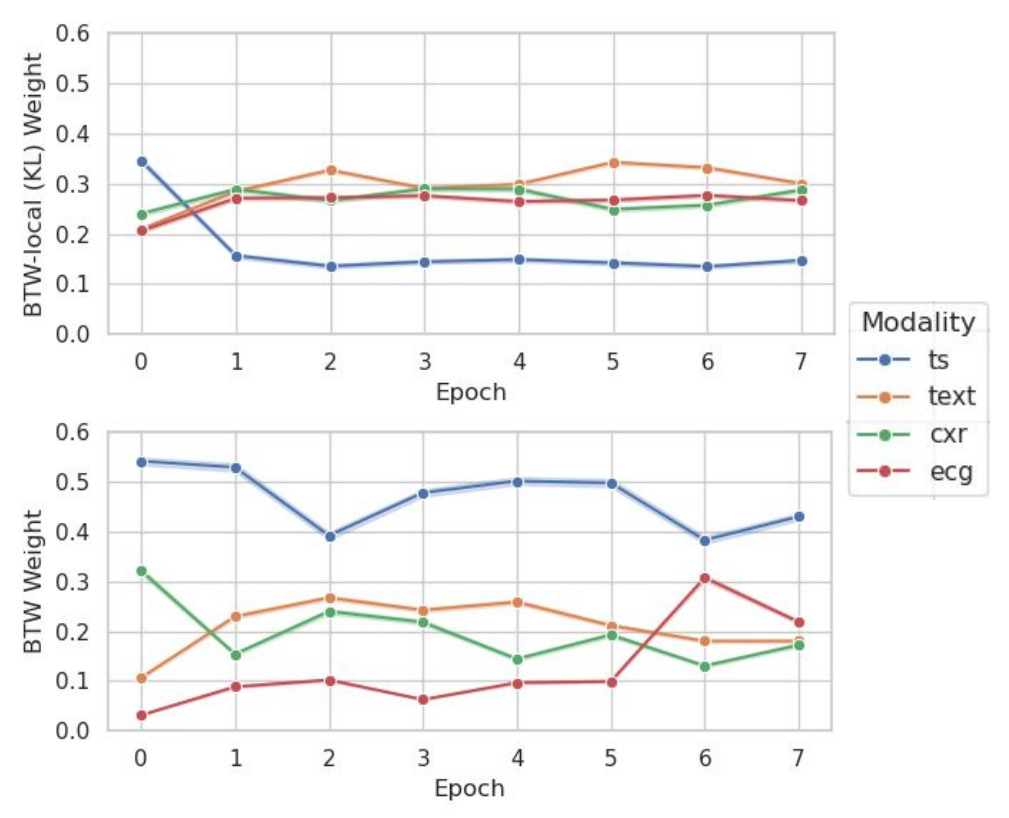}
    \caption{The evolution of modality weights across eight training epochs on MIMIC-IV dataset. The upper panel shows the BTW and the lower panel shows BTW-local (KL) weights.  BTW effectively balances the dominant TS modality from overy emphasized the uniqueness.}
    \label{fig:mimic_weights}
\end{figure}
\begin{figure*}[t]
  \centering
  \includegraphics[width=0.95\textwidth]{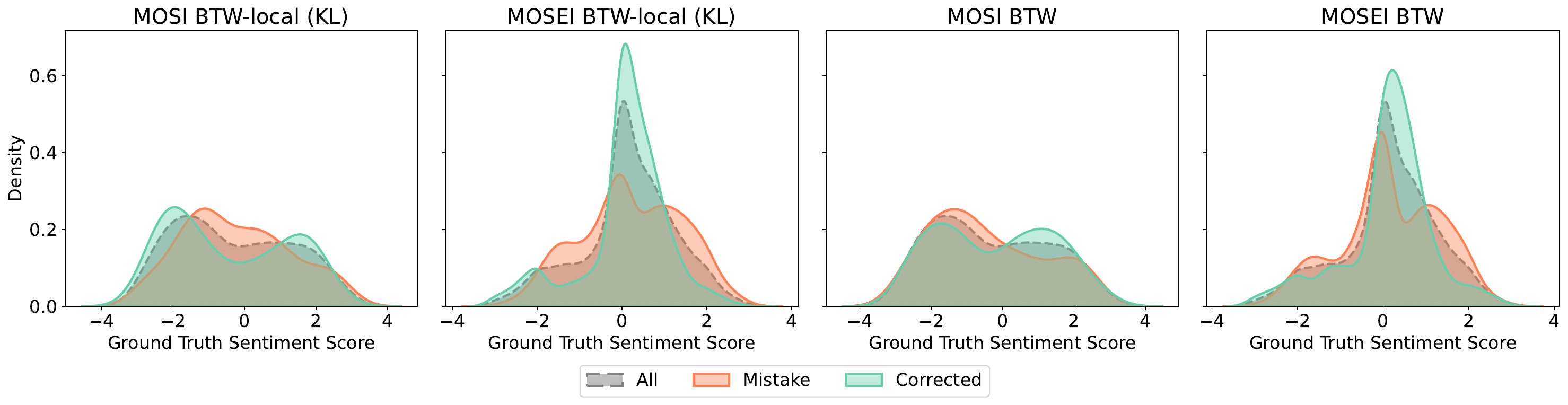}
  \caption{Kernel density plots of ground-truth sentiment scores for the test set, showing the distributions of all samples, predictions corrected by two variations of weighting schemes, and new mistakes introduced, across both MOSI and MOSEI datasets. 
  Both schemes tend to correct errors in high-density regions of the score distribution, with BTW especially concentrating corrections near neutral sentiment.}
  \label{fig:cmu-kl-combined-plot}
\end{figure*}
\begin{figure}[t]
  \centering
  \begin{subfigure}[ht]{0.48\textwidth}
    \centering
    \includegraphics[width=\textwidth]{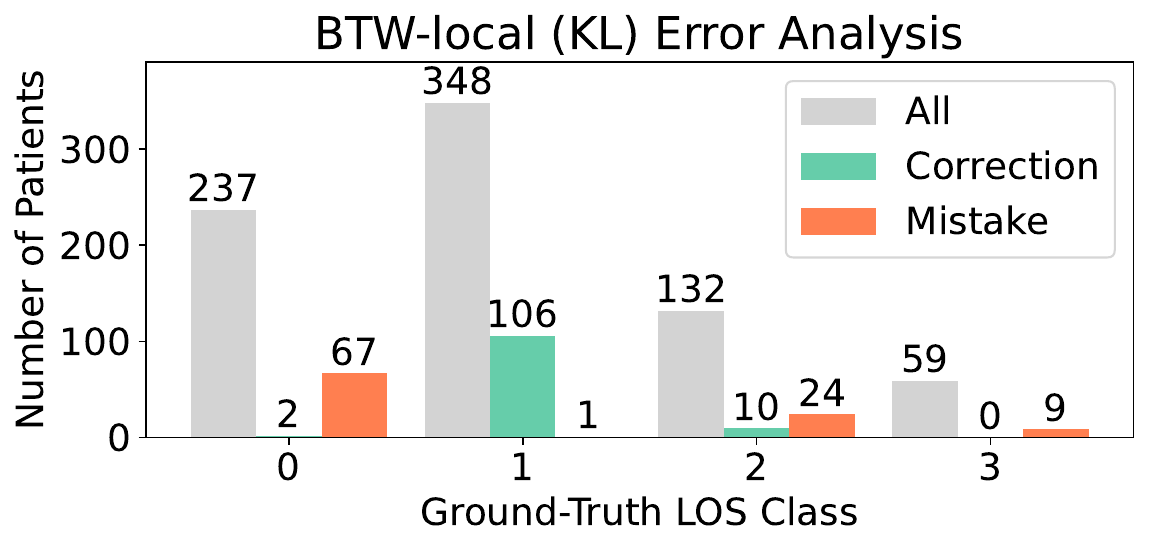}
  \end{subfigure}%
  \hfill
  \begin{subfigure}[ht]{0.48\textwidth}
    \centering
    \includegraphics[width=\textwidth]{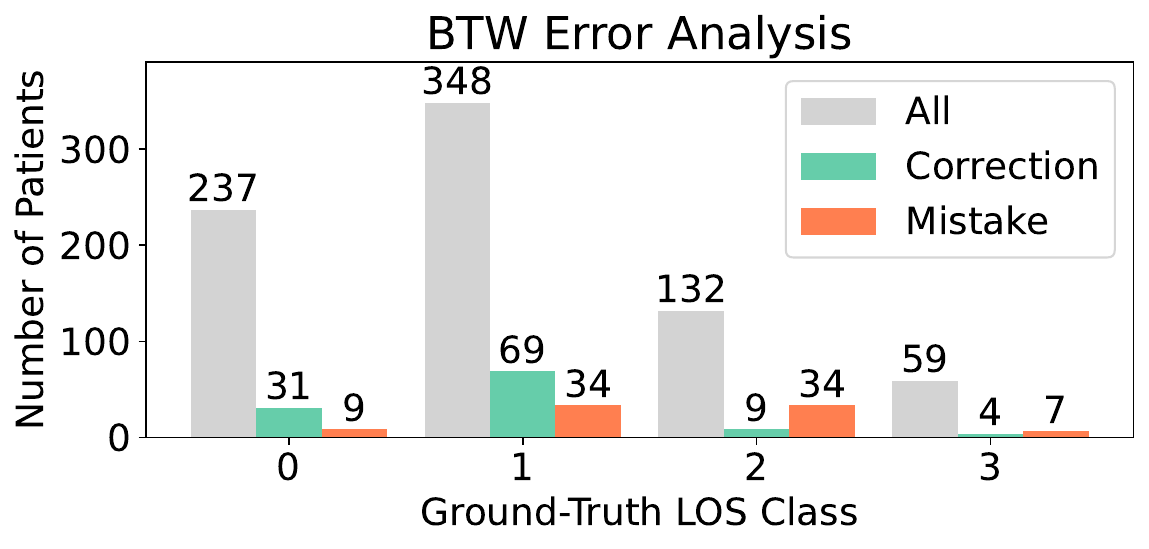}
  \end{subfigure}
  \caption{Visualization of test split counts of all instances, corrections, and mistakes over the ground-truth classes for the LOS classification task. BTW distributes corrections more evenly across all classes, notably improving performance in class 0.}
  \label{fig:mimic-error-plot}
\end{figure}
To better understand how our \our framework stabilizes variance in heterogeneous datasets, we analyze modality-specific weight trajectories across training epochs on MIMIC-IV (Figure~\ref{fig:mimic_weights}). 

In the BTW-local (KL) case, weights quickly stabilize after the first epoch. Notably, the weights for time-series modality (TS) drop sharply and remain relatively stable thereafter, indicating that the multimodal prediction for TS closely matches what can be achieved with TS alone. This observation aligns with our expectation that TS is inherently easier for the model to capture.
However, the high KL divergence weights for text, imaging (CXR), and ECG suggest that these modalities provide distinct, potentially noisy information that requires careful reconciliation during multimodal fusion.

Incorporating modality-level MI weights reshapes the distribution, consistently emphasizing the globally informative TS while balancing distinct contributions from other modalities, significantly improving accuracy and F1 scores (Table~\ref{tab:mimic-results}). Without weights, the model treats modalities equally, whereas BTW-local (KL) weights overly emphasize the uniqueness of TS. Integrating MI provides a necessary global perspective, balancing the strong predictive power of TS with the unique contributions of other modalities, resulting in a more robust multimodal fusion.

\subsection{Encoder Sensitivity}

To assess sensitivity to the language encoder, we replace the baseline BERT encoder with DeBERTa while keeping all other components identical, and compare against ITHP and other baselines (Table~\ref{tab:deberta-results}). BTW-local provides the strongest regression metrics, while BTW leads multi-class classification. On MOSI, BTW-local attains MAE=0.691; BTW achieves the best multiclass accuracy and high binary accuracy when including the zero-threshold (82.80\%). Similar trends hold on MOSEI, where BTW-local yields competitive MAE and BTW attains leading Acc-7/Acc-5. 

These results show that our framework captures fine-grained sentiment distinctions better than ITHP, which compresses information toward a dominant modality. In contrast, BTW balances all modalities using instance-level KL divergence and modality-level mutual information, preserving diverse signals and improving complex multi-class performance. This demonstrates that our weighting approach yields consistent improvements that are not dependent on the choice of encoder.

\begin{table*}[ht]
\centering
\small
\setlength{\tabcolsep}{3pt} 
\caption{
  Performance comparison on CMU-MOSI and CMU-MOSEI using the DeBERTa text encoder. 
  Results are compared against baseline models and the ITHP model. 
  Bold numbers indicate the best performance and underlined numbers indicate the second-best. 
  All results are averaged over three runs with different random seeds.
  The `Acc-2` and `Weighted-F1` columns show `include-zero / non-zero` results.
}
\label{tab:deberta-results}
\begin{tabular}{@{}l l cccccc@{}}
\toprule
& \textbf{Method} & \textbf{MAE$\downarrow$} & \textbf{Corr$\uparrow$} & \textbf{Acc-7$\uparrow$} & \textbf{Acc-5$\uparrow$} & \textbf{Acc-2$\uparrow$} & \textbf{Weighted-F1$\uparrow$} \\
\midrule
\multirow{5}{*}{\rotatebox[origin=c]{90}{\textbf{MOSI}}}
& ITHP      & 0.719\scriptsize{$\pm$0.004} & \textbf{0.840}\scriptsize{$\pm$0.006} & 43.75\scriptsize{$\pm$0.75} & 54.26\scriptsize{$\pm$1.22} & \textbf{85.69}\scriptsize{$\pm$1.02} / \textbf{87.33}\scriptsize{$\pm$1.00} & \textbf{85.66}\scriptsize{$\pm$1.05} / \textbf{87.34}\scriptsize{$\pm$1.02} \\
& MMIM      & 0.713\scriptsize{$\pm$0.007} & 0.799\scriptsize{$\pm$0.005} & 45.00\scriptsize{$\pm$0.99} & 52.72\scriptsize{$\pm$1.00} & 82.75\scriptsize{$\pm$0.17} / 84.93\scriptsize{$\pm$0.37} & 82.64\scriptsize{$\pm$0.23} / 84.77\scriptsize{$\pm$0.49} \\
& MoE       & 0.701\scriptsize{$\pm$0.011} & 0.793\scriptsize{$\pm$0.013} & 44.70\scriptsize{$\pm$1.09} & 52.48\scriptsize{$\pm$1.19} & 82.46\scriptsize{$\pm$0.89} / 84.80\scriptsize{$\pm$0.93} & 82.30\scriptsize{$\pm$0.96} / 84.73\scriptsize{$\pm$0.97} \\
\cmidrule{2-8}
& BTW-local & \underline{0.696}\scriptsize{$\pm$0.006} & \underline{0.801}\scriptsize{$\pm$0.005} & 45.14\scriptsize{$\pm$2.09} & 52.63\scriptsize{$\pm$1.65} & 83.53\scriptsize{$\pm$0.67} / 85.77\scriptsize{$\pm$1.07} & 83.38\scriptsize{$\pm$0.66} / 85.70\scriptsize{$\pm$1.07} \\
& BTW       & \textbf{0.691}\scriptsize{$\pm$0.005} & 0.790\scriptsize{$\pm$0.013} & \textbf{46.55}\scriptsize{$\pm$1.38} & \textbf{54.86}\scriptsize{$\pm$1.35} & 82.80\scriptsize{$\pm$0.50} / 85.37\scriptsize{$\pm$0.16} & 82.70\scriptsize{$\pm$0.46} / 85.04\scriptsize{$\pm$0.62} \\
\midrule[0.5pt]
\midrule[0.5pt]
\multirow{5}{*}{\rotatebox[origin=c]{90}{\textbf{MOSEI}}}
& ITHP      & 0.599\scriptsize{$\pm$0.004} & \textbf{0.782}\scriptsize{$\pm$0.003} & 48.30\scriptsize{$\pm$0.28} & 52.32\scriptsize{$\pm$0.32} & 80.67\scriptsize{$\pm$0.39} / \textbf{85.66}\scriptsize{$\pm$0.32} & 81.36\scriptsize{$\pm$0.34} / \textbf{85.75}\scriptsize{$\pm$0.29} \\
& MMIM      & 0.566\scriptsize{$\pm$0.014} & 0.732\scriptsize{$\pm$0.020} & 51.83\scriptsize{$\pm$0.67} & 53.11\scriptsize{$\pm$0.80} & 80.58\scriptsize{$\pm$1.86} / 83.34\scriptsize{$\pm$1.04} & 80.77\scriptsize{$\pm$1.35} / 83.03\scriptsize{$\pm$1.42} \\
& MoE       & 0.559\scriptsize{$\pm$0.006} & 0.742\scriptsize{$\pm$0.018} & 52.32\scriptsize{$\pm$0.46} & 53.72\scriptsize{$\pm$0.78} & 81.16\scriptsize{$\pm$2.62} / 83.90\scriptsize{$\pm$1.64} & 81.39\scriptsize{$\pm$1.95} / 83.68\scriptsize{$\pm$2.04} \\
\cmidrule{2-8}
& BTW-local & \textbf{0.540}\scriptsize{$\pm$0.005} & 0.757\scriptsize{$\pm$0.005} & \textbf{53.89}\scriptsize{$\pm$0.19} & \textbf{55.63}\scriptsize{$\pm$0.21} & 81.75\scriptsize{$\pm$1.23} / 85.11\scriptsize{$\pm$0.32} & 82.10\scriptsize{$\pm$0.92} / 84.97\scriptsize{$\pm$0.50} \\
& BTW       & \underline{0.545}\scriptsize{$\pm$0.007} & 0.756\scriptsize{$\pm$0.004} & \underline{53.35}\scriptsize{$\pm$0.27} & \underline{55.04}\scriptsize{$\pm$0.16} & \textbf{82.10}\scriptsize{$\pm$2.96} / 84.71\scriptsize{$\pm$0.45} & \textbf{82.34}\scriptsize{$\pm$2.47} / 83.48\scriptsize{$\pm$1.85} \\
\bottomrule
\end{tabular}
\end{table*}

\subsection{Smoothing Factor ($\alpha_t$)} \label{smooth}

We analyze the impact of EMA smoothing factor $\alpha_t\in{0.3,0.5,0.7}$ on BTW-local and BTW on MOSI/MOSEI. As shown in Table~\ref{tab:smoothing-factor-ablation}, lowering (0.3) or raising (0.7) improves some metrics while reducing others, so we set the default value to $\alpha_t=0.5$, which consistently offers the best and most stable trade-off across datasets and metrics.

\begin{table*}[ht]
\centering
\small
\setlength{\tabcolsep}{3pt} 
\caption{Ablation study on the smoothing factor ($\alpha_t$) for BTW-local (KL) and BTW methods on the MOSI and MOSEI datasets. The default value of 0.5 consistently provides the best or most stable performance. All results are averaged over three runs.}
\label{tab:smoothing-factor-ablation}
\begin{tabular}{@{}l l c cccccc@{}}
\toprule
& \textbf{Method} & \textbf{$\alpha_t$} & \textbf{MAE$\downarrow$} & \textbf{Corr$\uparrow$} & \textbf{Acc-7$\uparrow$} & \textbf{Acc-5$\uparrow$} & \textbf{Acc-2$\uparrow$} & \textbf{Weighted-F1$\uparrow$} \\
\midrule
\multirow{6}{*}{\rotatebox[origin=c]{90}{\textbf{MOSI}}} 
& \multirow{3}{*}{BTW-local} 
  & 0.3 & 0.740\scriptsize{$\pm$0.016} & 0.774\scriptsize{$\pm$0.008} & 45.48\scriptsize{$\pm$1.055} & 52.53\scriptsize{$\pm$1.752} & 81.88\scriptsize{$\pm$0.670}/83.94\scriptsize{$\pm$0.919} & 81.74\scriptsize{$\pm$0.688}/83.88\scriptsize{$\pm$0.952} \\
& & \textbf{0.5} & \textbf{0.714}\scriptsize{$\pm$0.011} & \textbf{0.786}\scriptsize{$\pm$0.005} & \textbf{46.40}\scriptsize{$\pm$3.227} & \textbf{53.26}\scriptsize{$\pm$3.312} & \textbf{82.46}\scriptsize{$\pm$0.970}/\textbf{84.55}\scriptsize{$\pm$0.919} & \textbf{82.33}\scriptsize{$\pm$0.961}/\textbf{84.50}\scriptsize{$\pm$0.883} \\
& & 0.7 & 0.758\scriptsize{$\pm$0.006} & 0.766\scriptsize{$\pm$0.003} & 43.64\scriptsize{$\pm$1.091} & 50.19\scriptsize{$\pm$1.701} & 81.29\scriptsize{$\pm$0.734}/83.54\scriptsize{$\pm$0.531} & 80.94\scriptsize{$\pm$0.960}/82.65\scriptsize{$\pm$1.473} \\
\cmidrule{2-9}
& \multirow{3}{*}{BTW} 
  & 0.3 & 0.739\scriptsize{$\pm$0.021} & 0.771\scriptsize{$\pm$0.011} & 45.73\scriptsize{$\pm$1.097} & 52.67\scriptsize{$\pm$1.604} & 81.54\scriptsize{$\pm$1.091}/83.79\scriptsize{$\pm$1.424} & 81.35\scriptsize{$\pm$0.865}/83.54\scriptsize{$\pm$1.427} \\
& & \textbf{0.5} & \textbf{0.716}\scriptsize{$\pm$0.008} & \textbf{0.781}\scriptsize{$\pm$0.009} & \textbf{47.52}\scriptsize{$\pm$0.773} & \textbf{54.28}\scriptsize{$\pm$1.431} & \textbf{82.75}\scriptsize{$\pm$1.173}/\textbf{84.35}\scriptsize{$\pm$0.836} & \textbf{82.68}\scriptsize{$\pm$1.260}/\textbf{84.34}\scriptsize{$\pm$0.910} \\
& & 0.7 & 0.744\scriptsize{$\pm$0.013} & 0.770\scriptsize{$\pm$0.006} & 44.95\scriptsize{$\pm$1.244} & 52.19\scriptsize{$\pm$1.197} & 81.97\scriptsize{$\pm$0.656}/84.60\scriptsize{$\pm$0.703} & 81.64\scriptsize{$\pm$0.606}/84.39\scriptsize{$\pm$0.650} \\
\midrule[0.5pt]
\midrule[0.5pt]
\multirow{6}{*}{\rotatebox[origin=c]{90}{\textbf{MOSEI}}} 
& \multirow{3}{*}{BTW-local} 
  & 0.3 & 0.574\scriptsize{$\pm$0.007} & 0.721\scriptsize{$\pm$0.004} & 52.14\scriptsize{$\pm$0.440} & 53.73\scriptsize{$\pm$0.188} & 82.83\scriptsize{$\pm$0.710}/82.90\scriptsize{$\pm$0.613} & 82.49\scriptsize{$\pm$0.490}/82.26\scriptsize{$\pm$0.765} \\
& & \textbf{0.5} & \textbf{0.566}\scriptsize{$\pm$0.008} & \textbf{0.727}\scriptsize{$\pm$0.006} & \textbf{52.32}\scriptsize{$\pm$0.757} & \textbf{53.85}\scriptsize{$\pm$0.732} & \textbf{83.02}\scriptsize{$\pm$1.098}/\textbf{83.60}\scriptsize{$\pm$1.869} & \textbf{82.81}\scriptsize{$\pm$0.983}/\textbf{83.07}\scriptsize{$\pm$2.288} \\
& & 0.7 & 0.563\scriptsize{$\pm$0.003} & 0.729\scriptsize{$\pm$0.003} & 52.69\scriptsize{$\pm$0.428} & 54.25\scriptsize{$\pm$0.359} & 81.21\scriptsize{$\pm$4.540}/82.64\scriptsize{$\pm$0.538} & 82.17\scriptsize{$\pm$1.989}/82.17\scriptsize{$\pm$0.386} \\
\cmidrule{2-9}
& \multirow{3}{*}{BTW} 
  & 0.3 & 0.574\scriptsize{$\pm$0.002} & 0.718\scriptsize{$\pm$0.005} & 52.07\scriptsize{$\pm$0.363} & 53.69\scriptsize{$\pm$0.174} & 81.56\scriptsize{$\pm$1.712}/82.43\scriptsize{$\pm$0.771} & 81.41\scriptsize{$\pm$1.642}/81.50\scriptsize{$\pm$1.525} \\
& & \textbf{0.5} & \textbf{0.573}\scriptsize{$\pm$0.007} & \textbf{0.722}\scriptsize{$\pm$0.003} & \textbf{52.62}\scriptsize{$\pm$0.737} & \textbf{54.15}\scriptsize{$\pm$0.867} & \textbf{81.97}\scriptsize{$\pm$2.356}/\textbf{81.92}\scriptsize{$\pm$0.961} & \textbf{81.50}\scriptsize{$\pm$1.976}/\textbf{81.22}\scriptsize{$\pm$1.360} \\
& & 0.7 & 0.577\scriptsize{$\pm$0.008} & 0.716\scriptsize{$\pm$0.006} & 52.03\scriptsize{$\pm$0.397} & 53.12\scriptsize{$\pm$0.817} & 82.41\scriptsize{$\pm$1.025}/82.55\scriptsize{$\pm$0.882} & 82.21\scriptsize{$\pm$0.856}/82.04\scriptsize{$\pm$0.760} \\
\bottomrule
\end{tabular}
\end{table*}

\subsection{Case Study: Error Analysis} 

To further validate how our proposed bi-level weighting framework improves multimodal predictions, we conduct an in-depth case study examining the instances where our methods correct baseline errors or introduce new mistakes. Kernel density estimates of ground-truth scores for all instances, corrections, and new mistakes are plotted for sentiment analysis regression (Figure~\ref{fig:cmu-kl-combined-plot};  complementary scatter plots in Appendix~\ref{sec:AppendixC}) and LOS classification tasks (Figure~\ref{fig:mimic-error-plot}).

When applied to both tasks, both weighting schemes consistently correct predictions in regions with high data density in the regression task, as higher density implies greater agreement and stronger confidence in corrections via KL divergence. For sentiment analysis, incorporating global MI leads to corrections concentrated around neutral sentiment scores in MOSI and MOSEI. For LOS classification, the results highlight a critical trade-off and the strength of our bi-level approach. While BTW-local (KL) focuses its corrections on the high-density majority class (106 corrections in class 1), it performs poorly on the most challenging minority class (only 2 corrections in class 0). In contrast, the full BTW framework uses global MI to re-balance its focus. It strategically sacrifices some corrections on the majority class to make significant gains in the most difficult classes, most notably increasing corrections in class 0 from 2 to 31. This ability to improve performance on minority classes by resolving uncertainty and ambiguity in classification boundaries is critical for building a robust and clinically useful model. 

While the BTW-local (KL) weights efficiently reduce variance in densely populated sentiment regions, global MI effectively targets the ambiguity regions, underscoring the complementary strengths of the bi-level weights.

\section{Conclusion}
This paper introduces a bi-level, non-parametric weighting framework that advances multimodal learning beyond two modalities by addressing prediction variance and modality interaction explainability. By integrating instance-level KL divergence with modality-level mutual information, the method adaptively calibrates modality contributions without introducing additional trainable parameters. The framework's value lies not only in quantitative improvements but also in its efficiency as a non-parametric, plug-and-play module that enhances existing architectures. Extensive experiments on diverse benchmarks and both regression and classification tasks demonstrate how the framework stabilizes variance in high-density regions and resolves ambiguity by leveraging globally informative modalities, ultimately facilitating more robust and transparent multimodal models.

\section*{Acknowledgement}
Our work is sponsored by NSF \#2442253, NAIRR Pilot with PSC Neocortex and NCSA Delta, Commonwealth Cyber Initiative, Children’s National Hospital, Fralin Biomedical Research Institute (Virginia Tech), Sanghani Center for AI and Data Analytics (Virginia Tech), Virginia Tech Innovation Campus, and generous gifts from Nivida, Cisco, and the Amazon + Virginia Tech Center for Efficient and Robust Machine Learning.

\clearpage
\section*{Limitations}
While the \our framework offers improved interpretability and performance across diverse multimodal tasks, there are some limitations that might obstacle further generalization and effectiveness. First, the information theory-based weights rely on the assumption that unimodal predictions provide informative and well-calibrated distributions. In cases where modalities have low quality or completeness, the resulting weights may introduce noise rather than stabilize variance. Future work could explore uncertainty-aware regularization or confidence-based gating to downweight unreliable unimodal predictions.

First, the information theory-based weights rely on the assumption that unimodal predictions provide informative and well-calibrated distributions. In cases where modalities have low quality or completeness, the resulting weights may introduce noise rather than stabilize variance. Future work could explore uncertainty-aware regularization or confidence-based gating to downweight unreliable unimodal predictions.

Second, the BTW framework, in its current form, is demonstrated on an MoE architecture. Its core requirement is the ability to obtain separate predictions from each unimodal path as well as a joint multimodal prediction. Therefore, it is directly applicable to various late-fusion or hybrid-fusion architectures but is not suited for pure early-fusion models where raw features are concatenated at the input layer, preventing the generation of distinct unimodal outputs from the fused representation.

Third, our method inherits the zero-embedding strategy from the FuseMoE backbone~\cite{han2024fusemoe} for handling missing modalities. As shown in our ablation, information-theoretic metrics like mutual information fail to provide meaningful signals when modalities are absent, due to the degenerate nature of zero vectors. Risks might arise if modality imputation is inaccurate or missing data are handled improperly, leading to unreliable or misleading model predictions. This suggests that imputing missing modality embeddings with synthetically generated representations could offer a more coherent and informative approximation, preserving the multimodal distributional structure.

Finally, while our method is task-agnostic for regression and classification tasks, supervision through labels is required for the current framework. This supervised assumption limits generalizability to unsupervised or self-supervised settings. Future research could explore proxy objectives such as contrastive similarity or mutual predictability to extend this framework to representation learning.

\section*{Ethics Statement}
All datasets used in this research are publicly available for research use, under the terms and licenses specified by their creators. MIMIC-IV is released under the PhysioNet Credentialed Health Data Use Agreement~\footnote{\url{https://physionet.org/content/mimiciv/2.2/}}. No proprietary data or restricted-access resources were used. We do not display raw excerpts from any dataset in this paper. We do not attempt to identify or deanonymize users in the
data in any way during our research.


\bibliography{anthology,custom}

\begin{thebibliography}{37}
\expandafter\ifx\csname natexlab\endcsname\relax\def\natexlab#1{#1}\fi

\bibitem[{Akbari et~al.(2023)Akbari, Kondratyuk, Cui, Hornung, Wang, and Adam}]{akbari2023alternating}
Hassan Akbari, Dan Kondratyuk, Yin Cui, Rachel Hornung, Huisheng Wang, and Hartwig Adam. 2023.
\newblock Alternating gradient descent and mixture-of-experts for integrated multimodal perception.
\newblock \emph{Advances in Neural Information Processing Systems}, 36:79142--79154.

\bibitem[{Bagher~Zadeh et~al.(2018)Bagher~Zadeh, Liang, Poria, Cambria, and Morency}]{bagher-zadeh-etal-2018-multimodal}
AmirAli Bagher~Zadeh, Paul~Pu Liang, Soujanya Poria, Erik Cambria, and Louis-Philippe Morency. 2018.
\newblock \href {https://doi.org/10.18653/v1/P18-1208} {Multimodal language analysis in the wild: {CMU}-{MOSEI} dataset and interpretable dynamic fusion graph}.
\newblock In \emph{Proceedings of the 56th Annual Meeting of the Association for Computational Linguistics (Volume 1: Long Papers)}, pages 2236--2246, Melbourne, Australia. Association for Computational Linguistics.

\bibitem[{Cao et~al.(2023)Cao, Sun, Zhu, and Hu}]{cao2023multi}
Bing Cao, Yiming Sun, Pengfei Zhu, and Qinghua Hu. 2023.
\newblock Multi-modal gated mixture of local-to-global experts for dynamic image fusion.
\newblock In \emph{Proceedings of the IEEE/CVF international conference on computer vision}, pages 23555--23564.

\bibitem[{Chen et~al.(2023)Chen, Liang, Wang, Zadeh, and Morency}]{chen2023multiviz}
Xiang Chen, Paul~Pu Liang, Xuan Wang, Amir Zadeh, and Louis-Philippe Morency. 2023.
\newblock Multiviz: Towards visualizing and understanding multimodal models.
\newblock In \emph{Proceedings of the 61st Annual Meeting of the Association for Computational Linguistics (ACL)}.

\bibitem[{Dai et~al.(2022)Dai, Tang, Liu, Tan, Zhou, Wang, Feng, Zhang, Hu, and Shi}]{dai2022one}
Yong Dai, Duyu Tang, Liangxin Liu, Minghuan Tan, Cong Zhou, Jingquan Wang, Zhangyin Feng, Fan Zhang, Xueyu Hu, and Shuming Shi. 2022.
\newblock One model, multiple modalities: A sparsely activated approach for text, sound, image, video and code.
\newblock \emph{arXiv preprint arXiv:2205.06126}.

\bibitem[{Fang et~al.(2024)Fang, Wu, Zhang, Huang, Zeng, Xing, Walsh, and Yang}]{fang2024dynamic}
Yingying Fang, Shuang Wu, Sheng Zhang, Chaoyan Huang, Tieyong Zeng, Xiaodan Xing, Simon Walsh, and Guang Yang. 2024.
\newblock Dynamic multimodal information bottleneck for multimodality classification.
\newblock In \emph{Proceedings of the IEEE/CVF Winter Conference on Applications of Computer Vision}, pages 7696--7706.

\bibitem[{Fedus et~al.(2022)Fedus, Zoph, and Shazeer}]{fedus2022switch}
William Fedus, Barret Zoph, and Noam Shazeer. 2022.
\newblock Switch transformers: Scaling to trillion parameter models with simple and efficient sparsity.
\newblock \emph{Journal of Machine Learning Research}, 23(120):1--39.

\bibitem[{Feng et~al.(2022)Feng, Zhang, Yu, Fang, Li, Chen, Lu, Liu, Yin, Feng, Sun, Tian, Wu, and Wang}]{Feng2022ERNIEViLG2IA}
Zhida Feng, Zhenyu Zhang, Xintong Yu, Yewei Fang, Lanxin Li, Xuyi Chen, Yuxiang Lu, Jiaxiang Liu, Weichong Yin, Shi Feng, Yu~Sun, Hao Tian, Hua Wu, and Haifeng Wang. 2022.
\newblock \href {https://api.semanticscholar.org/CorpusId:253157690} {Ernie-vilg 2.0: Improving text-to-image diffusion model with knowledge-enhanced mixture-of-denoising-experts}.
\newblock \emph{2023 IEEE/CVF Conference on Computer Vision and Pattern Recognition (CVPR)}, pages 10135--10145.

\bibitem[{Hager et~al.(2024)Hager, Jungmann, Holland, Bhagat, Hubrecht, Knauer, Vielhauer, Makowski, Braren, Kaissis et~al.}]{hager2024evaluation}
Paul Hager, Friederike Jungmann, Robbie Holland, Kunal Bhagat, Inga Hubrecht, Manuel Knauer, Jakob Vielhauer, Marcus Makowski, Rickmer Braren, Georgios Kaissis, et~al. 2024.
\newblock Evaluation and mitigation of the limitations of large language models in clinical decision-making.
\newblock \emph{Nature medicine}, 30(9):2613--2622.

\bibitem[{Han et~al.(2021)Han, Chen, and Poria}]{han2021improving}
Wei Han, Hui Chen, and Soujanya Poria. 2021.
\newblock Improving multimodal fusion with hierarchical mutual information maximization for multimodal sentiment analysis.
\newblock In \emph{Proceedings of the 2021 Conference on Empirical Methods in Natural Language Processing}, pages 9180--9192.

\bibitem[{Han et~al.(2024)Han, Nguyen, Harris, Ho, and Saria}]{han2024fusemoe}
Xing Han, Huy Nguyen, Carl Harris, Nhat Ho, and Suchi Saria. 2024.
\newblock \href {https://proceedings.neurips.cc/paper_files/paper/2024/file/7d62a85ebfed2f680eb5544beae93191-Paper-Conference.pdf} {Fusemoe: Mixture-of-experts transformers for fleximodal fusion}.
\newblock In \emph{Advances in Neural Information Processing Systems}, volume~37, pages 67850--67900. Curran Associates, Inc.

\bibitem[{He et~al.(2024)He, Cheng, Balasubramaniam, Tsai, and Zhao}]{he2024efficient}
Yifei He, Runxiang Cheng, Gargi Balasubramaniam, Yao-Hung~Hubert Tsai, and Han Zhao. 2024.
\newblock Efficient modality selection in multimodal learning.
\newblock \emph{Journal of Machine Learning Research}, 25(47):1--39.

\bibitem[{Johnson et~al.(2023)Johnson, Bulgarelli, Shen, Gayles, Shammout, Horng, Pollard, Hao, Moody, Gow et~al.}]{johnson2023mimic}
Alistair~EW Johnson, Lucas Bulgarelli, Lu~Shen, Alvin Gayles, Ayad Shammout, Steven Horng, Tom~J Pollard, Sicheng Hao, Benjamin Moody, Brian Gow, et~al. 2023.
\newblock Mimic-iv, a freely accessible electronic health record dataset.
\newblock \emph{Scientific data}, 10(1):1.

\bibitem[{Kontras et~al.(2024)Kontras, Strypsteen, Chatzichristos, Liang, Blaschko, and De~Vos}]{kontras2024multimodal}
Konstantinos Kontras, Thomas Strypsteen, Christos Chatzichristos, Paul~Pu Liang, Matthew Blaschko, and Maarten De~Vos. 2024.
\newblock Multimodal fusion balancing through game-theoretic regularization.
\newblock \emph{arXiv preprint arXiv:2411.07335}.

\bibitem[{Kullback(1997)}]{kullback1997information}
Solomon Kullback. 1997.
\newblock \emph{Information theory and statistics}.
\newblock Courier Corporation.

\bibitem[{Li et~al.(2025)Li, Jiang, Hu, Wang, Zhong, Luo, Ma, and Zhang}]{li2025uni}
Yunxin Li, Shenyuan Jiang, Baotian Hu, Longyue Wang, Wanqi Zhong, Wenhan Luo, Lin Ma, and Min Zhang. 2025.
\newblock Uni-moe: Scaling unified multimodal llms with mixture of experts.
\newblock \emph{IEEE Transactions on Pattern Analysis and Machine Intelligence}.

\bibitem[{Liang et~al.(2023)Liang, Cheng, Fan, Ling, Nie, Chen, Deng, Allen, Auerbach, Mahmood et~al.}]{liang2023quantifying}
Paul~Pu Liang, Yun Cheng, Xiang Fan, Chun~Kai Ling, Suzanne Nie, Richard Chen, Zihao Deng, Nicholas Allen, Randy Auerbach, Faisal Mahmood, et~al. 2023.
\newblock Quantifying \& modeling multimodal interactions: An information decomposition framework.
\newblock \emph{Advances in Neural Information Processing Systems}, 36:27351--27393.

\bibitem[{Lin et~al.(2024{\natexlab{a}})Lin, Tang, Ye, Cui, Zhu, Jin, Huang, Zhang, Pang, Ning et~al.}]{lin2024moe}
Bin Lin, Zhenyu Tang, Yang Ye, Jiaxi Cui, Bin Zhu, Peng Jin, Jinfa Huang, Junwu Zhang, Yatian Pang, Munan Ning, et~al. 2024{\natexlab{a}}.
\newblock Moe-llava: Mixture of experts for large vision-language models.
\newblock \emph{arXiv preprint arXiv:2401.15947}.

\bibitem[{Lin and Hu(2023)}]{lin2023missmodal}
Ronghao Lin and Haifeng Hu. 2023.
\newblock Missmodal: Increasing robustness to missing modality in multimodal sentiment analysis.
\newblock \emph{Transactions of the Association for Computational Linguistics}, 11:1686--1702.

\bibitem[{Lin and Hu(2024)}]{lin2024adapt}
Ronghao Lin and Haifeng Hu. 2024.
\newblock Adapt and explore: Multimodal mixup for representation learning.
\newblock \emph{Information Fusion}, 105:102216.

\bibitem[{Lin et~al.(2024{\natexlab{b}})Lin, Shrivastava, Luo, Iyer, Lewis, Ghosh, Zettlemoyer, and Aghajanyan}]{lin2024moma}
Xi~Victoria Lin, Akshat Shrivastava, Liang Luo, Srinivasan Iyer, Mike Lewis, Gargi Ghosh, Luke Zettlemoyer, and Armen Aghajanyan. 2024{\natexlab{b}}.
\newblock Moma: Efficient early-fusion pre-training with mixture of modality-aware experts.
\newblock \emph{arXiv preprint arXiv:2407.21770}.

\bibitem[{Lyu et~al.(2022)Lyu, Liang, Deng, Salakhutdinov, and Morency}]{lyu2022dime}
Yiwei Lyu, Paul~Pu Liang, Zihao Deng, Ruslan Salakhutdinov, and Louis-Philippe Morency. 2022.
\newblock Dime: Fine-grained interpretations of multimodal models via disentangled local explanations.
\newblock In \emph{Proceedings of the 2022 AAAI/ACM Conference on AI, Ethics, and Society}, pages 455--467.

\bibitem[{Mustafa et~al.(2022)Mustafa, Riquelme, Puigcerver, Jenatton, and Houlsby}]{mustafa2022multimodal}
Basil Mustafa, Carlos Riquelme, Joan Puigcerver, Rodolphe Jenatton, and Neil Houlsby. 2022.
\newblock Multimodal contrastive learning with limoe: the language-image mixture of experts.
\newblock \emph{Advances in Neural Information Processing Systems}, 35:9564--9576.

\bibitem[{Poklukar et~al.(2022)Poklukar, Vasco, Yin, Melo, Paiva, and Kragic}]{poklukar2022geometric}
Petra Poklukar, Miguel Vasco, Hang Yin, Francisco~S Melo, Ana Paiva, and Danica Kragic. 2022.
\newblock Geometric multimodal contrastive representation learning.
\newblock In \emph{International Conference on Machine Learning}, pages 17782--17800. PMLR.

\bibitem[{Shannon(1948)}]{shannon1948mathematical}
Claude~E Shannon. 1948.
\newblock A mathematical theory of communication.
\newblock \emph{The Bell system technical journal}, 27(3):379--423.

\bibitem[{Shazeer et~al.(2017)Shazeer, Mirhoseini, Maziarz, Davis, Le, Hinton, and Dean}]{shazeer2017outrageously}
Noam Shazeer, Azalia Mirhoseini, Krzysztof Maziarz, Andy Davis, Quoc Le, Geoffrey Hinton, and Jeff Dean. 2017.
\newblock Outrageously large neural networks: The sparsely-gated mixture-of-experts layer.
\newblock \emph{arXiv preprint arXiv:1701.06538}.

\bibitem[{Soenksen et~al.(2022)Soenksen, Ma, Zeng, Boussioux, Villalobos~Carballo, Na, Wiberg, Li, Fuentes, and Bertsimas}]{soenksen2022integrated}
Luis~R Soenksen, Yu~Ma, Cynthia Zeng, Leonard Boussioux, Kimberly Villalobos~Carballo, Liangyuan Na, Holly~M Wiberg, Michael~L Li, Ignacio Fuentes, and Dimitris Bertsimas. 2022.
\newblock Integrated multimodal artificial intelligence framework for healthcare applications.
\newblock \emph{NPJ digital medicine}, 5(1):149.

\bibitem[{Tsai et~al.(2019)Tsai, Bai, Liang, Kolter, Morency, and Salakhutdinov}]{tsai-etal-2019-multimodal}
Yao-Hung~Hubert Tsai, Shaojie Bai, Paul~Pu Liang, J.~Zico Kolter, Louis-Philippe Morency, and Ruslan Salakhutdinov. 2019.
\newblock \href {https://doi.org/10.18653/v1/P19-1656} {Multimodal transformer for unaligned multimodal language sequences}.
\newblock In \emph{Proceedings of the 57th Annual Meeting of the Association for Computational Linguistics}, pages 6558--6569, Florence, Italy. Association for Computational Linguistics.

\bibitem[{van Aken et~al.(2021)van Aken, Papaioannou, Mayrdorfer, Budde, Gers, and Löser}]{vanAken2021}
Betty van Aken, Jens-Michalis Papaioannou, Manuel Mayrdorfer, Klemens Budde, Felix~A. Gers, and Alexander Löser. 2021.
\newblock \href {https://www.aclweb.org/anthology/2021.eacl-main.75/} {Clinical outcome prediction from admission notes using self-supervised knowledge integration}.
\newblock In \emph{Proceedings of the 16th Conference of the European Chapter of the Association for Computational Linguistics: Main Volume, {EACL} 2021, Online, April 19 - 23, 2021}, pages 881--893. Association for Computational Linguistics.

\bibitem[{Williams and Beer(2010)}]{williams2010nonnegative}
Paul~L Williams and Randall~D Beer. 2010.
\newblock Nonnegative decomposition of multivariate information.
\newblock \emph{arXiv preprint arXiv:1004.2515}.

\bibitem[{Wu et~al.(2023)Wu, Dai, Qin, Liu, Lin, Cao, and Sui}]{wu2023denoising}
Shaoxiang Wu, Damai Dai, Ziwei Qin, Tianyu Liu, Binghuai Lin, Yunbo Cao, and Zhifang Sui. 2023.
\newblock Denoising bottleneck with mutual information maximization for video multimodal fusion.
\newblock In \emph{Proceedings of the 61st Annual Meeting of the Association for Computational Linguistics (Volume 1: Long Papers)}, pages 2231--2243.

\bibitem[{Xue and Marculescu(2023)}]{xue2023dynamic}
Zihui Xue and Radu Marculescu. 2023.
\newblock Dynamic multimodal fusion.
\newblock In \emph{Proceedings of the IEEE/CVF Conference on Computer Vision and Pattern Recognition}, pages 2575--2584.

\bibitem[{Yun et~al.(2024)Yun, Choi, Peng, Wu, Bao, Zhang, Xin, Long, and Chen}]{NEURIPS2024_b2f2af54}
Sukwon Yun, Inyoung Choi, Jie Peng, Yangfan Wu, Jingxuan Bao, Qiyiwen Zhang, Jiayi Xin, Qi~Long, and Tianlong Chen. 2024.
\newblock \href {https://proceedings.neurips.cc/paper_files/paper/2024/file/b2f2af5403042b1344f4e93b35fb67d9-Paper-Conference.pdf} {Flex-moe: Modeling arbitrary modality combination via the flexible mixture-of-experts}.
\newblock In \emph{Advances in Neural Information Processing Systems}, volume~37, pages 98782--98805. Curran Associates, Inc.

\bibitem[{Zadeh et~al.(2016)Zadeh, Zellers, Pincus, and Morency}]{zadeh2016mosi}
Amir Zadeh, Rowan Zellers, Eli Pincus, and Louis-Philippe Morency. 2016.
\newblock Mosi: multimodal corpus of sentiment intensity and subjectivity analysis in online opinion videos.
\newblock \emph{arXiv preprint arXiv:1606.06259}.

\bibitem[{Zadeh et~al.(2018)Zadeh, Liang, Poria, Cambria, and Morency}]{zadeh2018multimodal}
AmirAli~Bagher Zadeh, Paul~Pu Liang, Soujanya Poria, Erik Cambria, and Louis-Philippe Morency. 2018.
\newblock Multimodal language analysis in the wild: Cmu-mosei dataset and interpretable dynamic fusion graph.
\newblock In \emph{Proceedings of the 56th Annual Meeting of the Association for Computational Linguistics (Volume 1: Long Papers)}, pages 2236--2246.

\bibitem[{Zhang et~al.(2023)Zhang, Doughty, and Snoek}]{zhang2023learning}
Yunhua Zhang, Hazel Doughty, and Cees Snoek. 2023.
\newblock Learning unseen modality interaction.
\newblock \emph{Advances in Neural Information Processing Systems}, 36:54716--54726.

\bibitem[{Zhao et~al.(2024)Zhao, Wang, Tan, and Wang}]{zhao2024tgmoe}
Xueliang Zhao, Mingyang Wang, Yingchun Tan, and Xianjie Wang. 2024.
\newblock Tgmoe: A text guided mixture-of-experts model for multimodal sentiment analysis.
\newblock \emph{International Journal of Advanced Computer Science \& Applications}, 15(8).

\end{thebibliography}
\bibliographystyle{acl_natbib}

\clearpage
\appendix

\section{MIMIC-IV with Missing Modalities}
\label{app:missing}
In real-world, the more modalities included will result in more likely and challenging missing modality problem. We conducted additional experiments using the MIMIC-IV dataset including stays with missing modalities as shown in Table~\ref{tab:mimic-missing-results}. While the BTW-local (KL) variant achieves slightly better Macro-F1 (39.67\%), we observe a performance drop after applying the full \our weights, particularly when incorporating modality-level MI. These results indicate instance-level weights provide robust and stable performance even when modalities are absent. In contrast, the modality-level MI becomes unreliable, since the missing modality contributes no information, fooled the weights wrongly favors modalities that are consistently present regardless of their quality in informativeness. This bias degrades the variance stabilization ability for the bi-level weights, revealing the inherent sensitivity of MI-based alignment to incomplete real-world data.

\begin{table}[htbp]
\centering
\small
\caption{
Performance on the MIMIC-IV dataset with randomly missing modalities. 
BTW-local (KL) and BTW represent variants of our bi-level weighting framework. 
All metrics are averaged over three random seeds. Standard deviations are shown in \scriptsize{$\pm$}.
}
\label{tab:mimic-missing-results}
\begin{tabular}{lccc}
\toprule
\textbf{Method} & \textbf{Accuracy} & \textbf{Macro-F1} & \textbf{Weighted-F1} \\
\midrule
MulT     & 46.67\scriptsize{$\pm$1.53} & 38.33\scriptsize{$\pm$3.06} & 45.67\scriptsize{$\pm$1.53} \\
FuseMoE      & 44.00\scriptsize{$\pm$1.00} & 38.67\scriptsize{$\pm$1.53} & 44.00\scriptsize{$\pm$2.00} \\
\midrule
\makecell{BTW-\\local (KL)}           & 45.67\scriptsize{$\pm$0.58} & 39.67\scriptsize{$\pm$1.15} & 45.00\scriptsize{$\pm$0.00} \\
BTW              & 44.00\scriptsize{$\pm$1.41} & 35.50\scriptsize{$\pm$0.71} & 42.50\scriptsize{$\pm$0.71} \\
\bottomrule
\end{tabular}
\end{table}

\section{3-Modalities Performance}
\label{app:3mod}
This section presents additional results evaluating the robustness of our bi-level weighting framework against FuseMoE~\cite{han2024fusemoe} under various modality ablation scenarios on MIMIC-IV \cite{johnson2023mimic}. We compare model performance when each of the three auxiliary modalities (ECG, CXR, Text) is removed individually, under both no missing (Table~\ref{tab:mimic_full_results} and full datasets Table~\ref{tab:mimic_miss_results}).

\begin{table}[ht]
\centering
\small
\caption{Performance comparison across weighting strategies on MIMIC without missing modalities}
\begin{tabular}{llccc}
\toprule
\textbf{Method} & \textbf{Modality} & \textbf{Acc} & \textbf{\makecell{Macro \\ F1}} & \textbf{\makecell{Weighted \\ F1}} \\
\hline
\multirow{4}{*}{FuseMoE} 
& w/o ECG         & 46 & 41 & 46 \\
& w/o Text        & 43 & 42 & 41 \\
& w/o CXR         & 48 & 37 & 47 \\
\hline
\hline
\multirow{4}{*}{\makecell{BTW-local\\(KL)}} 
& w/o ECG         & 47 & 40   & 46 \\
& w/o Text        & 42 & 33.5 & 40 \\
& w/o CXR         & 44 & 38.5 & 44 \\
\hline
\hline
\multirow{4}{*}{BTW} 
& w/o ECG         & 45 & 37 & 44 \\
& w/o Text        & 45 & 17 & 29 \\
& w/o CXR         & 46 & 36 & 43 \\
\bottomrule
\end{tabular}
\label{tab:mimic_full_results}
\end{table}

\begin{table}[ht]
\centering
\small
\caption{Performance comparison across weighting strategies on full MIMIC dataset}
\begin{tabular}{llccc}
\toprule
\textbf{Method} & \textbf{Modality} & \textbf{Acc} & \textbf{\makecell{Macro \\ F1}} & \textbf{\makecell{Weighted \\ F1}} \\
\hline
\multirow{4}{*}{FuseMoE} 
& w/o ECG         & 45 & 38 & 43 \\
& w/o Text        & 43 & 35 & 40 \\
& w/o CXR         & 47 & 38 & 46 \\
\hline
\hline
\multirow{4}{*}{\makecell{BTW-local\\(KL)}} 
& w/o ECG         & 46 & 36   & 44 \\
& w/o Text        & 44 & 30   & 40 \\
& w/o CXR         & 44 & 36   & 42 \\
\hline
\hline
\multirow{4}{*}{BTW} 
& w/o ECG         & 46 & 34   & 43 \\
& w/o Text        & 44 & 25   & 36 \\
& w/o CXR         & 46 & 37   & 45 \\
\bottomrule
\end{tabular}
\label{tab:mimic_miss_results}
\end{table}

\section{Proof of Estimation of Conditional Variance} \label{app:proof}
This section formally derives the instance-level KL divergence used in our weighting framework for regression tasks. We provide a justification based on the Law of Total Variance, show the unbiasedness of residuals, and present the closed-form expression for Gaussian KL divergence between unimodal and multimodal predictions.

\subsection{Law of Total Variance}
\begin{equation}
  \operatorname{Var}(Y)
  = \operatorname{Var}\bigl(\mathrm{E}[Y \mid X]\bigr)
  + \mathrm{E}\bigl[\operatorname{Var}(Y \mid X)\bigr]
  \label{eq:total-var-appendix}
\end{equation}

\begin{equation}
  \operatorname{Var}\bigl(\mathrm{E}[Y \mid X_i]\bigr)
  = \operatorname{Var}(\mu_i) = 0
  \label{eq:var-mu-zero}
\end{equation}

\begin{equation}
  \operatorname{Var}(Y \mid X_i)
  = \operatorname{Var}(Y)
  = \mathrm{E}\bigl[(Y - \mu_i)^2\bigr]
  \label{eq:cond-var-appendix}
\end{equation}

\subsection{Unbiasedness of the Squared Residual}
\begin{equation}
  \mathrm{E}\bigl[(Y - \mu_i)^2 \mid X_i\bigr]
  = \operatorname{Var}(Y \mid X_i),
  \label{eq:unbiased-residual}
\end{equation}
so each empirical squared residual \((Y_i - \mu_i)^2\) is an unbiased estimate of the conditional variance.

\subsection{Gaussian KL–Divergence}
For two Gaussians \(p = \mathcal{N}(\mu_p,\sigma_p)\) and \(q = \mathcal{N}(\mu_q,\sigma_q)\),
\begin{equation}
  D_\text{KL}(p\;\|\;q)
  = \log\!\frac{\sigma_q}{\sigma_p}
    + \frac{\sigma_p^2 + (\mu_p - \mu_q)^2}{2\,\sigma_q^2}
    - \tfrac12.
  \label{eq:gauss-kl}
\end{equation}
By setting
\[
  \mu_p = \mu_i,
  \quad
  \sigma_p^2 = (Y_i - \mu_i)^2,
  \]
\[
  \mu_q = \mu_i^{\mathrm{multi}},
  \quad
  \sigma_q = (Y_i - \mu_i^{\mathrm{multi}})^2,
\]
we obtain the instance–level weight
\begin{multline}
  w_i = D_\text{KL}\bigl(\mathcal{N}(\mu_i,\,(Y_i - \mu_i)^2)\;\|\; \\
          \mathcal{N}(\mu_i^{\mathrm{multi}},\,(Y_i - \mu_i^{\mathrm{multi}})^2\bigr).
\end{multline}

\section{Scatter plots for error analysis} \label{sec:AppendixC}
\begin{figure}[ht]
  \centering
  \begin{subfigure}[ht]{0.48\textwidth}
    \centering
    \includegraphics[width=\textwidth]{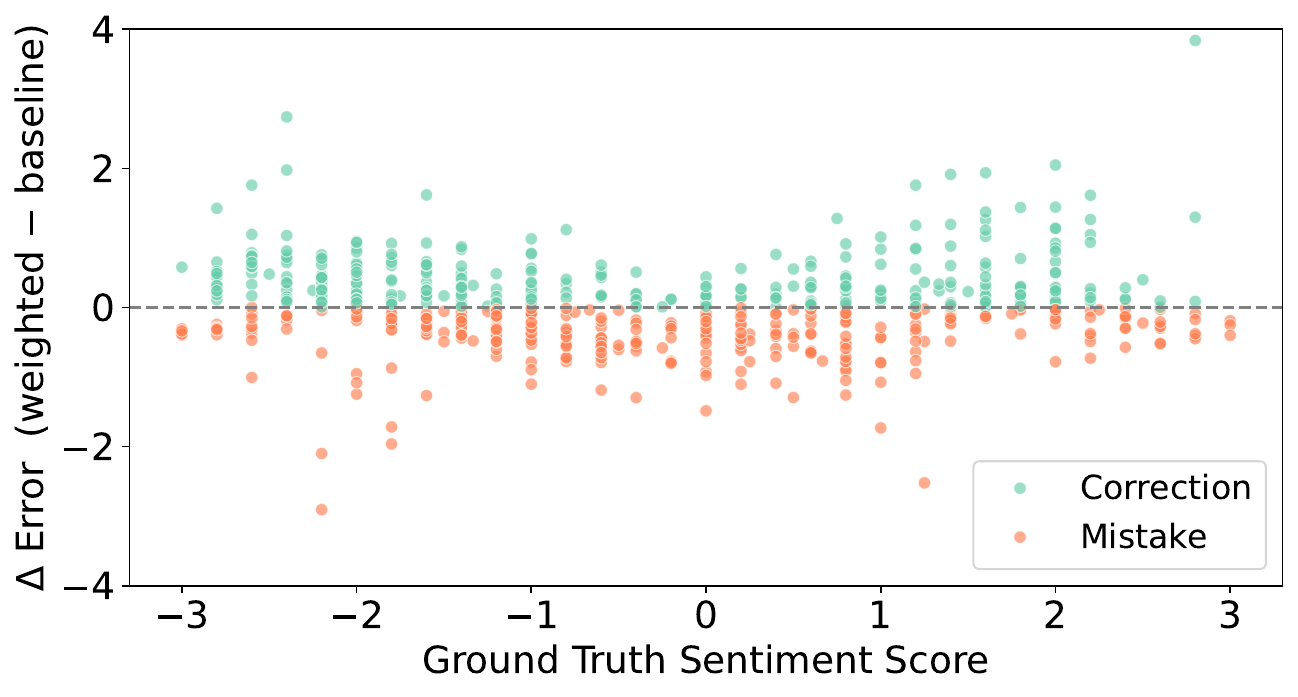}
    \caption{MOSI BTW-local (KL) Corrections and Mistakes Distribution}
  \end{subfigure}
  \hfill
  \begin{subfigure}[ht]{0.48\textwidth}
    \centering
    \includegraphics[width=\textwidth]{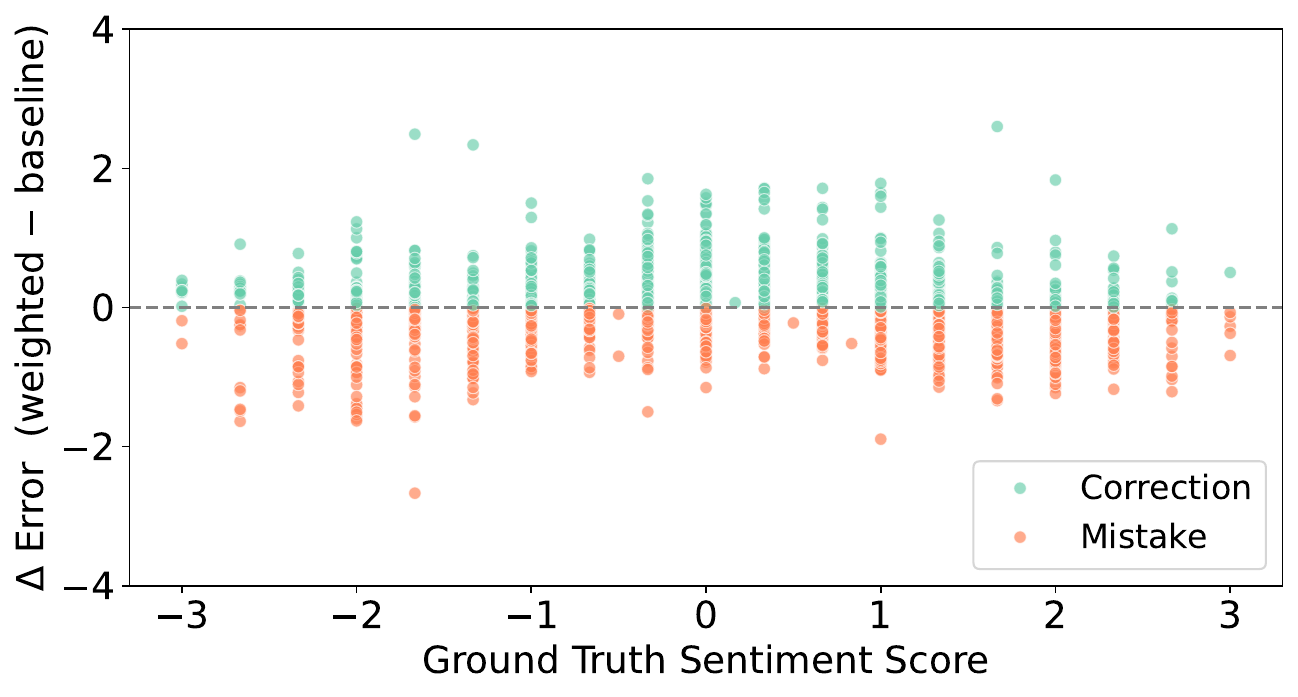}
    \caption{MOSEI BTW-local (KL) Corrections and Mistakes Distribution}
  \end{subfigure}

  \caption{Visualization of test split scatter plots, corrections density and mistakes density over the ground truth sentiment score for BTW-local (KL) experiments.}
  \label{fig:cmu-scatter-local}
\end{figure}

\begin{figure}[ht]
  \centering
  \begin{subfigure}[ht]{0.48\textwidth}
    \centering
    \includegraphics[width=\textwidth]{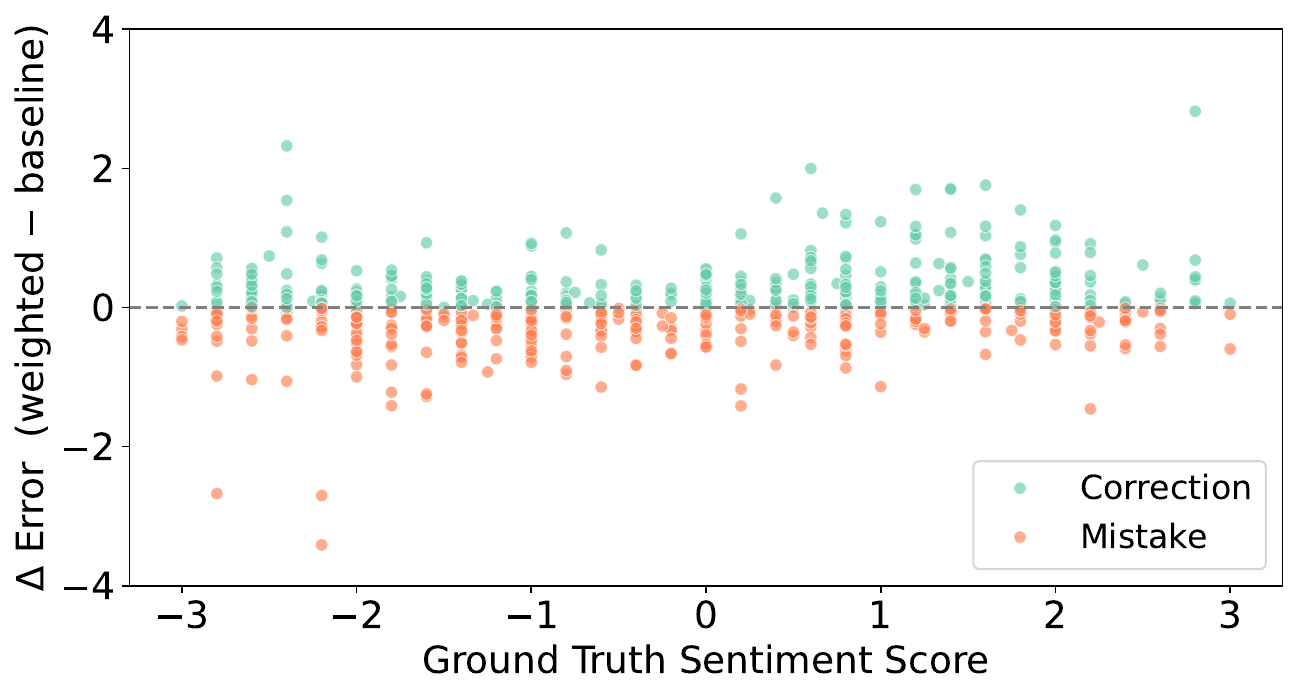}
    \caption{MOSI BTW Corrections and Mistakes Distribution}
  \end{subfigure}
  \hfill
  \begin{subfigure}[ht]{0.48\textwidth}
    \centering
    \includegraphics[width=\textwidth]{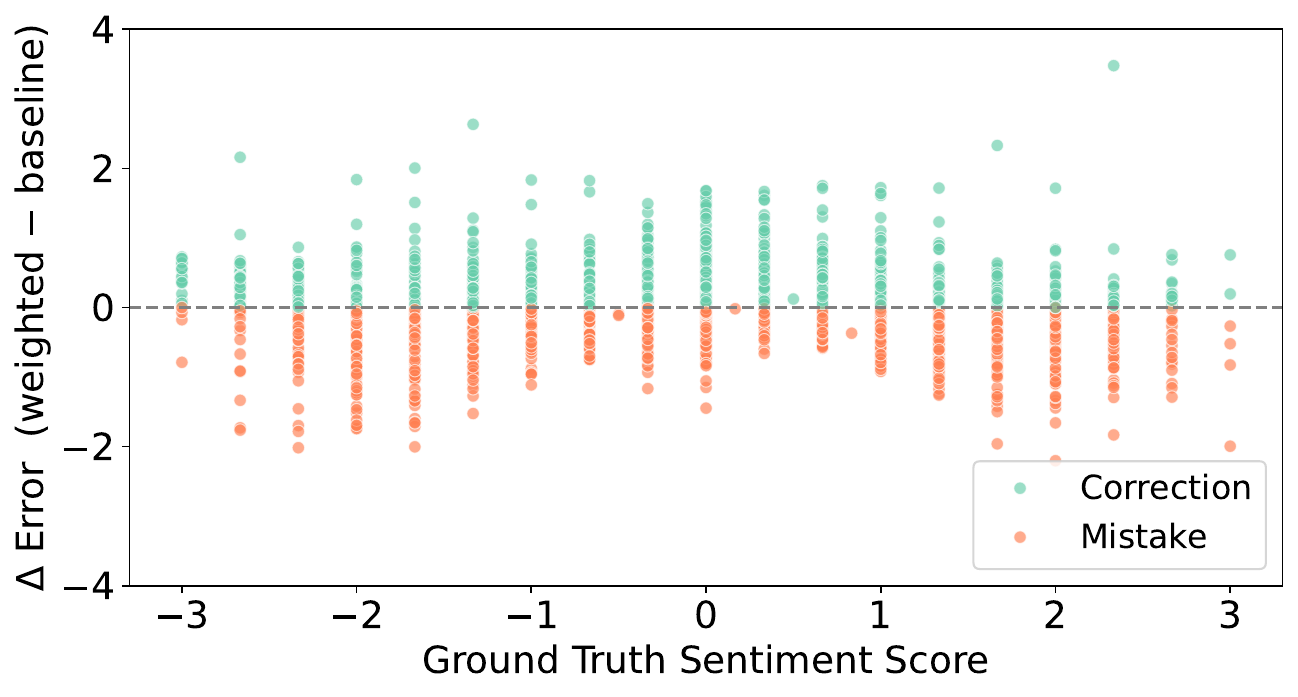}
    \caption{MOSEI BTW Corrections and Mistakes Distribution}
  \end{subfigure}
  \caption{Visualization of test split scatter plots corrections density and mistakes density over the ground truth sentiment score for BTW experiments.}
  \label{fig:cmu-scatter-BTW}
\end{figure}

We visualize instance-level corrections and mistakes for sentiment regression as shown in Figure~\ref{fig:cmu-scatter-local} and Figure~\ref{fig:cmu-scatter-BTW} for BTW-local (KL) and BTW scenario, respectively. Each point represents the change in prediction (corrected or error) relative to the true sentiment score, with the vertical axis indicating the magnitude and sign of these modifications. For MOSI, corrections using BTW-local (KL) tend to show larger error reductions in moderately positive and negative regions, whereas the BTW approach produces a denser band of modest corrections across the full range. In MOSEI, both schemes demonstrate more uniform correction magnitudes across sentiment values. In all cases, the majority of mistakes (orange) are smaller in magnitude and more evenly distributed, confirming the effectiveness of bi-level weights.

\section{Proof of Mutual Information} \label{app: MIproof}
Let \(\mathcal{X}\) be the whole dataset with \(N\) data points. Naturally \(P(X_i) = 1/N\) for all \(X_i \in \mathcal{X}\). Then we have:

\begin{align}
P(\hat{y}^{(m)}) &= \sum_{X^{(m)}_i \in \mathcal{X}} P(\hat{y}^{(m)}_i | X^{(m)}_i)\,P(X^{(m)}_i) \\
P(\hat{y}^{\text{multi}}) &= \sum_{X^{\text{multi}}_i \in \mathcal{X}} P(\hat{y}^{\text{multi}}_i | X^{\text{multi}}_i)\,P(X^{\text{multi}}_i)
\end{align}

And the joint probability is defined as:
\begin{multline}
P(\hat{y}^{(m)}, \hat{y}^\text{multi}) = \\ \sum_{X^{(m)}_i \in \mathcal{X}} P(\hat{y}^{(m)}_i | X^{(m)}_i) P(X^{(m)}_i|X^{\text{multi}}_i) \\
P(\hat{y}^{\text{multi}}_i | X^{\text{multi}}_i)P(X^{\text{multi}}_i)
\end{multline}

\section{Computation Resources, Time Cost, and Hyper-Parameters} \label{app: hyper}

\begin{table*}[h]
\centering
\small
\caption{Computational cost analysis. All times are averaged over three runs.}
\label{tab:time-cost}
\begin{tabular}{@{}l l r r r r r@{}}
\toprule
& \textbf{Method} & \textbf{Text Overhead} & \textbf{Audio Overhead} & \textbf{Video Overhead} & \textbf{Main Training Time} & \textbf{Total Est. Time} \\
\midrule
\multirow{3}{*}{\rotatebox[origin=c]{90}{\textbf{MOSI}}}
& Baseline        & N/A    & N/A    & N/A    & 6m 45s & 6m 45s \\
& BTW-local (KL)  & 6m 4s  & 2m 39s & 1m 7s  & 7m 35s & 17m 25s \\
& BTW (KL+MI)     & 6m 4s  & 2m 39s & 1m 7s  & 7m 55s & 17m 46s \\
\midrule
\multirow{3}{*}{\rotatebox[origin=c]{90}{\textbf{MOSEI}}}
& Baseline        & N/A    & N/A    & N/A    & 37m 36s & 37m 36s \\
& BTW-local (KL)  & 30m 4s & 22m 24s& 22m 20s& 43m 59s & 118m 47s \\
& BTW (KL+MI)     & 30m 4s & 22m 24s& 22m 20s& 45m 36s & 120m 24s \\
\bottomrule
\end{tabular}
\end{table*}

\begin{table}[htbp]
\centering
\small
\caption{Hyperparameters used for MoE module}
\label{tab:moe-hyper}
\begin{tabular}{lc}
\toprule
\textbf{Parameter Name} & \textbf{Value} \\
\midrule
Number of Experts & 16 \\
FFN hidden size & 512 \\
Top k & 2 \\
Router\_type & permod \\
Hidden activation function & GeLU \\
Number of MoE layers & 3 \\
\bottomrule
\end{tabular}
\end{table}

We summarize the computation resources and MoE-specific hyperparameters and packages used in our experiments. Hyper-parameter settings are aligned with those in prior work on LOS classification and sentiment analysis to ensure comparability and reproducibility.

\paragraph{Computation Resources}
All experiments were conducted on a single NVIDIA A40 GPU with 48GB memory (CUDA 12.4, driver version 550.67). For the CMU datasets, we follow the preprocessing steps from MMIM~\cite{han2021improving} and incorporate the MoE layer from FuseMoE~\cite{han2024fusemoe}. 

\paragraph{Time Cost}

Our framework introduces an initial, one-time overhead for pre-training the unimodal models, which is required to initialize the weighting process. The duration of this step is highly dependent on the size of the dataset. As shown in Table~\ref{tab:time-cost}, the per-epoch training time of our methods remains comparable to the baseline.

\paragraph{Hyper-Parameter Settings}
The hyper-parameters for MoE module are listed in Table~\ref{tab:moe-hyper}. Other hyper-parameters used for LOS classification are same as in \cite{han2024fusemoe}, and for sentiment analysis are same as in \cite{han2021improving}.

\paragraph{Implementation}
Our implementation used Python 3.8 with the following key libraries: PyTorch, NumPy, pandas, scikit-learn, and HuggingFace Transformers.



\end{document}